\documentclass[10pt, twocolumn, letterpaper]{article}

\newcommand{\final}{1}
\newcommand{\forArxiv}{1}

\usepackage[pagenumbers]{cvpr} 

\usepackage{graphicx}
\usepackage{amsmath}
\usepackage{amssymb}
\usepackage{booktabs}
\usepackage[abs]{overpic}
\usepackage{xcolor}
\usepackage{color}
\usepackage{ifthen}
\usepackage{float}
\usepackage{array}
\usepackage{multirow}

\newcommand{\nothing}[1]{}

\definecolor{bluegreen}{RGB}{16, 196, 178}
\definecolor{gold}{RGB}{196, 136, 16}

\definecolor{DeltaColor}{rgb}{0.039,0.73,0.71}
\definecolor{SetaColor}{rgb}{0.867, 0.0235, 0.376}
\definecolor{SigmaColor}{rgb}{0.98,0.45,0.0}
\definecolor{RedColor}{rgb}{0.8,0,0}
\definecolor{AlphaColor}{rgb}{0,0,0.8}
\definecolor{BetaColor}{rgb}{0.8,0,0.8}
\definecolor{GammaColor}{rgb}{0.5,0,0.7}
\definecolor{EpsilonColor}{rgb}{0.353,0.725,0.906}
\definecolor{TauColor}{rgb}{0.423,0.235,0.192}
\newcommand{\weikai}[1]{{\color{RedColor} Weikai: #1 $\qed$}}
\newcommand{\cheng}[1]{{\color{AlphaColor} Cheng: #1 $\qed$}}
\newcommand{\brandon}[1]{{\color{DeltaColor} Brandon: #1 $\qed$}}
\newcommand{\weiyang}[1]{{\color{GammaColor} Weiyang: #1 $\qed$}}

\newcommand{\warning}[1]{{\it\color{red} #1}}
\newcommand{\note}[1]{{\it\color{blue} #1}}

\definecolor{AudioColor}{rgb}{0.56,0.34,0.62}

\definecolor{DeadlineColor}{rgb}{0.9,0.4,0} 
\newcommand{\deadline}[1]{{\bf\color{DeadlineColor} ETA: #1}}

\definecolor{figred}{rgb}{1,0,0}
\definecolor{figgreen}{rgb}{0,0.6,0}
\definecolor{figblue}{rgb}{0,0,1}
\definecolor{figpink}{rgb}{1,0.63,0.63}

\newcolumntype{C}[1]{>{\centering}m{#1}}

\ifthenelse{\equal{\final}{1}}
{
\renewcommand{\weikai}[1]{}
\renewcommand{\cheng}[1]{}
\renewcommand{\weiyang}[1]{}
\renewcommand{\brandon}[1]{}
\renewcommand{\warning}[1]{}
\renewcommand{\note}[1]{}
\renewcommand{\deadline}[1]{}
}
{}

\newcounter{pccount}
\floatstyle{plain}

\newcommand{\filename}[1]{\url{#1}}
\newcommand{\foldername}[1]{\url{#1}}

\DeclareMathOperator*{\argmin}{argmin}         

\usepackage[pagebackref,breaklinks,colorlinks]{hyperref}

\def\cvprPaperID{*****}
\def\confYear{CVPR 2022}
\begin{document}
	%
	\title{3PSDF: Three-Pole Signed Distance Function for Learning Surfaces \\ with Arbitrary Topologies}
	
	\author{Weikai Chen \quad\quad Cheng Lin \quad\quad Weiyang Li \quad\quad Bo Yang\\
		Digital Content Technology Center, Tencent Games\\
		{\tt\small \{weikaichen,arnolin,kimonoli,brandonyang\}@tencent.com  }
	}

	
	\twocolumn[{%
\renewcommand\twocolumn[1][]{#1}%
\maketitle
\begin{center}
    \centering
    \vspace{-15pt}
  \includegraphics[width=\linewidth]{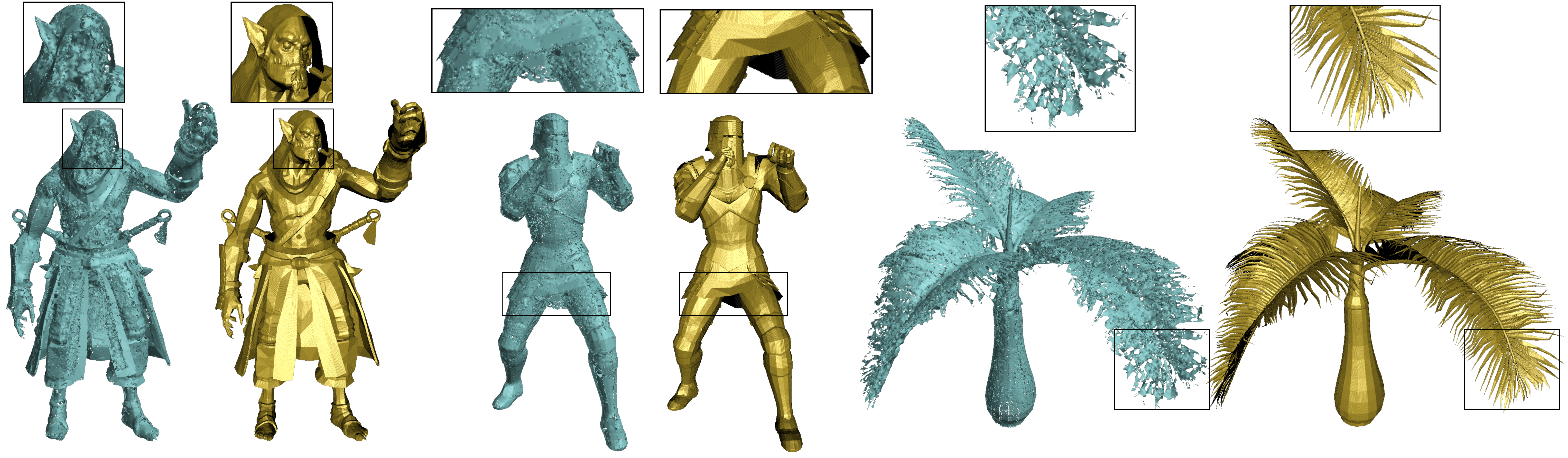}
    \vspace{-0.3in}
    \captionof{figure}{We show three groups of shape reconstruction results generated by NDF~\cite{chibane2020ndf} (in \textcolor{bluegreen}{cyan}) and our proposed 3PSDF (in \textcolor{gold}{gold}) respectively. Our method is able to faithfully reconstruct high-fidelity, intricate geometric details including both the closed and open surfaces, while NDF suffers from the meshing problems. Each NDF result is reconstructed from a dense point cloud containing 1 million points while ours are reconstructed using an equivalent resolution.}
    \label{fig:teaser}
\end{center}%
}]

	\begin{abstract}
Recent advances in learning 3D shapes using neural implicit functions have achieved impressive results by breaking the previous barrier of resolution and diversity for varying topologies. However, most of such approaches are limited to closed surfaces as they require the space to be divided into inside and outside. More recent works based on unsigned distance function have been proposed to handle complex geometry containing both the open and closed surfaces. Nonetheless, as their direct outputs are point clouds, robustly obtaining high-quality meshing results from discrete points remains an open question. We present a novel learnable implicit representation, called three-pole signed distance function (3PSDF), that can represent non-watertight 3D shapes with arbitrary topologies while supporting easy field-to-mesh conversion using the classic Marching Cubes algorithm. The key to our method is the introduction of a new sign, the NULL sign, in addition to the conventional in and out labels. The existence of the null sign could stop the formation of a closed isosurface derived from the bisector of the in/out regions. Further, we propose a dedicated learning framework to effectively learn 3PSDF without worrying about the vanishing gradient due to the null labels. Experimental results show that our approach outperforms the previous state-of-the-art methods in a wide range of benchmarks both quantitatively and qualitatively.

\end{abstract}

	\section{Introduction}
\label{sec:intro}

The choice of representation for 3D shapes and surfaces has been a central topic for effective 3D learning.
Various 3D representations, including mesh~\cite{wang2018pixel2mesh, gkioxari2019mesh}, voxels~\cite{tatarchenko2017octree, wang2017cnn}, and point cloud~\cite{qi2017pointnet, qi2017pointnet++}, have been extensively studied over the past years.
Recently, the advent of neural implicit functions (NIF)~\cite{huang2018deep,mescheder2019occupancy,park2019deepsdf,chen2019learning} has brought impressive advances to the state-of-the-art of learning-based 3D reconstruction and modeling.

Classic NIF approaches are built upon the signed distance function (SDF); they train a deep neural network to classify continuous 3D locations as inside or outside the surface via occupancy prediction or regressing the SDF.  
However, they can only model $\textit{closed}$ surfaces that support the in/out test for level surface extraction. 
Recent advances that leverage unsigned distance function (UDF)~\cite{chibane2020ndf,venkatesh2020dude,venkatesh2021deep} have made it possible to learn $\textit{open}$ surfaces from point clouds. 
But instantiating this field into an explicit mesh remains cumbersome and is prone to artifacts.
It requires the generation of dense point cloud and leveraging UDF's gradient field to iteratively push the points onto the target surface.
Such process is vulnerable to complex gradient landscape, e.g., parts with many details, and could easily get stuck at a local minima. In addition, reconstruction of mesh from UDF has to use the Ball Pivoting (BP) algorithm which has several drawbacks. 1) It is very sensitive to the input ball radius. A slightly larger or smaller radius would lead to an incomplete meshing result. 
2) It is prone to generate self-intersections and disconnected face patches with inconsistent normals even with surfaces of moderate complexity (see the clothing result in Figure~\ref{fig:recon}). 
3) The BP algorithm is very time-consuming especially dealing with dense point clouds. Finally, learning UDF becomes a regression task instead of classification like for SDF, making the training more difficult. We show in the closeups of Figure~\ref{fig:teaser} that NDF~\cite{chibane2020implicit} cannot reconstruct the face details of the first character even with 1 million sampling points.



\begin{figure*}[ht]
	\centering
	\includegraphics[width=1.0\textwidth]{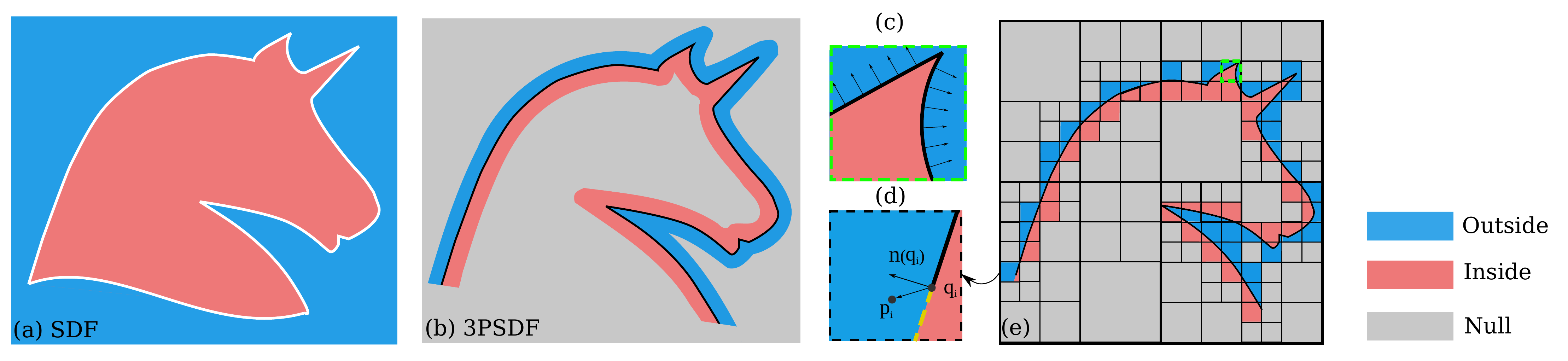}
	\caption{2D illustration of 3PSDF. (a) Conventional signed distance function (SDF) can only represent closed surface. (b) By introducing the null sign into SDF, 3PSDF can disable specified decision boundaries to cast arbitrary topologies that contain open surfaces. We propose practical framework for computing 3PSDF based on local cells ((c) and (d)). While 3PSDF may introduce approximation error (the yellow dash line in (d)) for open surface enclosed within a cell, the approximation error can be significantly reduced with finer space decomposition. We propose octree-based subdivision approach (e) to improve approximation performance with high computation efficiency.}
	\label{fig:overview}
\end{figure*}

We overcome the above limitations by presenting a new learnable implicit representation, called \textit{Three-Pole Signed Distance Function} (3PSDF), which is capable of representing highly intricate geometries containing both closed and open surfaces with high fidelity (see Figure~\ref{fig:teaser}). In addition, 3PSDF makes the learning an easy-to-train classification task, and is compatible with classic and efficient iso-surface extraction techniques, e.g. the Marching Cubes algorithm.
The key idea of our approach is the introduction of a direction-less sign, the \textit{NULL} sign, into the conventional binary-sided signed distance function. 
Points with null sign will be assigned with nan value, preventing the decision boundary to be formed between them and their neighbors. 
Therefore, by properly distributing the null signs over the space, we are able to cast surfaces with arbitrary topologies (see Figure~\ref{fig:overview}). 
Similar to previous works based on occupancy prediction~\cite{mescheder2019occupancy,chen2019learning}, we train a neural network to classify continuous points into 3 categories: \textit{inside}, \textit{outside}, and \textit{null}.
The resulting labels can be converted back to the 3PSDF using a simple mapping function to obtain meshing result.





We evaluate 3PSDF on three different tasks with gradually increased difficulty: shape reconstruction, point cloud completion and single-view reconstruction. 
3PSDF can consistently outperform the state-of-the-art methods over a wide range of benchmarks, including ShapeNet~\cite{chang2015shapenet}, MGN~\cite{bhatnagar2019mgn}, Maximo~\cite{Mixamo}, and 3D-Front~\cite{fu20203d}, both quantitatively and qualitatively.
We also conduct comparisons of field-to-mesh conversion time with NDF and analyze the impact of different resolutions and sampling strategies on our approach. Our contributions can be summarized as:


\begin{itemize}
    \item We present a new learnable 3D representation, 3PSDF, that can represent highly intricate shapes with both closed and open surfaces while being compatible with existing level surface extraction techniques.
    
    \item We propose a simple yet effective learning paradigm for 3PSDF that enables it to handle challenging task like single-view reconstruction.
    
    \item We obtain SOTA results on three applications across a wide range of benchmarks using 3PSDF.
\end{itemize}

\nothing{
\weikai{Test}
\cheng{Test}
\brandon{Test}
}
	\section{Related Work}
\label{sec:related_work}

\paragraph{Learning with explicit representations.} Explicit representations of 3D shapes are often well regularized and structured. Voxel based methods \cite{graham20183d, choy20194d, girdhar2016learning} are compatible with convolutional neural networks for learning; to reduce the high memory cost, octree-based partitions are adopted \cite{liu2020neural, tatarchenko2017octree, wang2017cnn}. However, inner parts of objects usually occupy a large portion of the voxels, leading to compromised 3D accuracy due to memory limitation. Mesh-based methods mostly deform a pre-defined mesh to approximate a given 3D shape \cite{dai2019scan2mesh, gkioxari2019mesh, wang2018pixel2mesh, pan2019deep}. One key limitation of such methods is the difficulty of changing mesh topologies, confining its 3D representation capability. Point clouds have achieved much attention recently \cite{shu20193d, qi2017pointnet++, valsesia2018learning, yang2019pointflow} due to its simplicity. Although such methods are convenient for shape analysis, generating 3D shapes with high precision remains difficult.

\paragraph{Implicit function learning.} With the development of deep learning, implicit representation of 3D shapes has achieved great progress in recent years \cite{chibane2020implicit, chen2019learning, duan2020curriculum, liu2019learning, natsume2019siclope, tang2021octfield}. A good example is signed distance field (SDF), which creates a continuous implicit field in 3D space \cite{park2019deepsdf, peng2020convolutional} where outside and inside points are denoted by positive and negative SDFs. Zero-isosurface, i.e., the object's surface, can be efficiently extracted by Marching Cubes~\cite{march}. Such representation supports infinite resolution and can simplify the SDF learning as a binary classification process\cite{mescheder2019occupancy}. However, SDF is only applicable to objects with closed surfaces.

To deal with open surfaces, unsigned distance field (UDF) \cite{chibane2020ndf} and deep unsigned distance embeddings \cite{venkatesh2020dude} are proposed. These methods use absolute distance to describe point position, and the zero-isosurface is extracted by Ball-Pivoting algorithm \cite{ballpivot}. However they have several major limitations: 1) learning UDF is a regression problem, harder than that in SDF; 2) ball pivoting \cite{ballpivot} is more computational expensive and less stable than Marching Cubes \cite{march}; 3) gradient vanishes on the surface, resulting in artifacts. Venkatesh et al \cite{venkatesh2021deep} proposed Closest Surface-Point (CSP) representation to prevent gradient vanishing and improve the speed. Zhao et al. \cite{zhao2021learning} proposed Anchor UDF to improve reconstruction accuracy. However, the first two limitations of UDF-type methods still remain.
	\section{Three-Pole Signed Distance Function}
\label{sec:method}

\subsection{Definition}
\label{sec:3psdf}
A watertight 3D shape can be implicitly represented by a signed distance function. Given a 3D query point $\mathbf{p} \in \mathbb{R}^3$, previous works apply deep neural networks to either predict the occupancy of $\mathbf{p}$ as $f(\mathbf{p}): \mathbb{R}^3 \mapsto [0, 1]$ \cite{huang2018deep} or directly regress SDF as $f(\mathbf{p}): \mathbb{R}^3 \mapsto \mathbb{R}$ \cite{park2019deepsdf,xu2019disn}. 
Our key observation is that the formation of closed surface is inevitable as long as both the positive and negative signs exist in the space (note that we do not consider space clipping where SDF is only computed in a limited bounding area).
To resolve this issue, we introduce the third direction-less pole -- the NULL sign into the field such that the ``curse" of closeness can be lifted: no iso-surfaces can be formed at the bisector of either positive/null or negative/null pairs. 
Therefore, the null sign acts as a surface eliminator that cancels out unwanted surfaces and thus can flexibly cast arbitrary topologies including those with open surfaces.

Formally, for a 3D point $\mathbf{p} \in \mathbb{R}^3$, we propose that in addition to a continuous signed distance, it can be also be mapped to null value: $\Psi(\mathbf{p}): \mathbb{R}^3 \mapsto \{\mathbb{R}, nan\}$. 
Hence, given an input surface $\mathbf{\mathcal{S}}$, we aim to learn such a mapping function $\Psi$ so that

\vspace{-2mm}
\begin{equation}
    \argmin_\Psi || \mathcal{S} - \mathcal{M}(\Psi(\mathbf{p})) ||, 
\end{equation}
\vspace{-2mm}

\noindent where $\mathcal{M}$ is the meshing operator that converts the resulting field into an explicit surface and $||\cdot||$ returns the surface to surface distance. Next, we will introduce how to compute the corresponding 3PSDF for a given shape.

\subsection{Field Computation}
\label{sec:compute_3psdf}

For non-watertight surface without closed boundaries, it is not possible to perform in/out test for a query point. 
Hence, we leverage the surface normal to determine the sign of the distance.
In particular, we decompose the 3D space into grid of local cells. 
As shown in Figure~\ref{fig:overview}, for each cell $\mathcal{C}_i$, if it does not contain any surface of interest, we set its enclosed space as null region and any sample point $\mathbf{p}_i$ that lies inside $\mathcal{C}_i$ has nan distance to the target surface $\mathcal{S}$:
\begin{equation}
    \Psi(\mathbf{p}_i, \mathcal{S}) = nan, \text{ if } \mathbf{p}_i \in \mathcal{C}_i \text{ and } \mathcal{C}_i \cap \mathcal{S} = \O
    \label{eqn:null}
\end{equation}

For a local cell $\mathcal{C}_i$ that encloses a surface patch $\mathcal{S}_i$, given a query point $\mathbf{p}_i \in \mathcal{C}_i$, we find $\mathbf{p}_i$'s closest point $\mathbf{q}_i$ on $\mathcal{S}_i$.
We set the surface normal at $\mathbf{q}_i$ as $\mathbf{n}(\mathbf{q}_i)$. 
If vector $\overrightarrow{\mathbf{q}_i\mathbf{p}_i}$ aligns with $\mathbf{n}(\mathbf{q}_i)$, i.e. $\mathbf{n}(\mathbf{q}_i) \cdot \overrightarrow{\mathbf{q}_i\mathbf{p}_i} \ge 0$, we set $\mathbf{p}_i$'s distance to the input surface $\mathcal{S}$ as positive; otherwise, it is negative.
The computation can be summarized as:

\vspace{-4mm}
\begin{equation}
\Psi(\mathbf{p}_i, \mathcal{S}_i) = 
    \begin{cases}
      \mathbf{d}(\mathbf{p}_i, \mathcal{S}_i)  & \text{if } \mathbf{n}(\mathbf{q}_i) \cdot \overrightarrow{\mathbf{q}_i\mathbf{p}_i} \ge 0, \\
      -\mathbf{d}(\mathbf{p}_i, \mathcal{S}_i) & \text{otherwise},\\
    \end{cases} 
    \label{eqn:surface}
\end{equation}

\noindent where $\mathbf{d}(\mathbf{p}, \mathcal{S}_i)$ returns the absolute distance between $\mathbf{p}_i$ and $\mathcal{S}_i$.
With finer decomposition of 3D space, cells containing geometry would only distribute around the surface of interest while the null cells would occupy the majority of the space. 
This differs a lot from the conventional signed distance field, where the entirety of the space is filled with distances of either positive or negative sign.
Our proposed 3PSDF better reflects the nature of 3D surface of any topology -- the high sparsity of surface occupancy.  

\paragraph{Surface approximation ability.} 
If an enclosed surface subdivides its hosting cell into several closed sub-regions, our implicit representation can faithfully approximate the original shape without loss of accuracy (Figure~\ref{fig:overview}(c)).
If a local cell contains protruding open surface(s), our approach is prone to generate elongated surface patch (Figure~\ref{fig:overview}(d)). 
However, such approximation error only happen locally and is limited to the size of the local cell. 
Hence, with a denser 3D decomposition, we can significantly reduce the approximation error. 
We provide additional experiments in Section~\ref{sec:exp_ablation} showing different reconstruction performance with respect to varying sampling resolutions.

\subsection{Learning Framework}
\label{sec:learn_3psdf}

Though the introduction of the null sign provides the flexibility of eliminating unwanted surface, the nan value prohibits computing meaningful gradient required for updating a deep neural network.
To resolve this issue, a straightforward way is to combine binary classification (nan v.s. non-nan) and regression, where the former generates a mask of the valid narrow band around the surface and the later regresses the surface within this narrow band.
While we experimentally validate that it is possible to learn 3PSDF via this approach, additional challenges would arise in aligning the narrow-band mask from binary classification and the regressed decision boundary from the regression branch. 
A misalignment of the two branches' results would lead to discontinuity in the final reconstruction.
Hence, we propose an alternative learning framework that formulates the learning of 3PSDF as a 3-way classification problem as elaborated below.
While we provide the method and results of the 3-way classification framework in the main paper, we provide detailed comparisons between the two learning methods in the supplemental materials.

Similar to the previous works on occupancy prediction~\cite{mescheder2019occupancy,chen2019learning}, the 3-way classification method proposes to approximate the target function (Equation~\ref{eqn:null} and \ref{eqn:surface}) with a neural network that infers per-point label: $\{in, out, null\}$. 
We represent the label semantics using discrete numbers without loss of generality.
Formally, we aim to learn a mapping function
 $o:\mathbb{R}^3 \mapsto \{0, 1, 2\},$
where the labels $\{0, 1, 2\}$ represent inside, outside, and null respectively.

When applying such a network for downstream tasks (e.g. 3D reconstruction) based on observation of the object (e.g. point cloud, image, etc.), the network must be conditioned on the input. 
Therefore, in addition to the coordinate of points $\mathbf{p} \in \mathbb{R}^3$, the network also consumes the observation of object $\mathbf{x} \in \mathcal{X}$ as input.
Specifically, such a mapping function can be parameterized by a neural network $\Phi_\theta$ that takes a pair $(\mathbf{p}, \mathbf{x})$ as input and outputs its 3-class label:

\vspace{-2mm}
\begin{equation}
    \Phi_\theta(\mathbf{p}, \mathbf{x}): \mathbb{R}^3 \times \mathcal{X} \mapsto \{0, 1, 2\}.
\end{equation}
\vspace{-6mm}

\paragraph{Training.}
To learn the parameters $\theta$ of the neural network $\Phi_\theta(\mathbf{p}, \mathbf{x})$, we train the network using batches of point samples. 
For the $i$-th sample in a training batch, we sample $N$ points $p_{ij} \in \mathbb{R}^3, j = 1, \dots, N$.
The mini-batch loss $\mathcal{L}_{\mathcal{B}}$ is:

\begin{equation}
\mathcal{L}_{\mathcal{B}} = \frac{1}{|\mathcal{B}|N} \sum^{|\mathcal{B}|}_{i=1}
\sum_{j=1}^{N} \mathcal{L}(\Phi_\theta(p_{ij}, x_i), y_{ij}),  
\end{equation}

\noindent where $\mathcal{L}(\cdot, \cdot)$ computes the cross-entropy loss, $x_i$ is the $i$-th observation of batch $\mathcal{B}$, $y_{ij}$ denotes the ground-truth label for point $p_{ij}$.

\paragraph{Octree-based subdivision.}
Since the computation of 3PSDF is done locally, to ensure a high reconstruction accuracy, it would be preferable not to include too many intricate geometric details and open surfaces in one cell. We propose an octree-based subdivision~\cite{wang2017cnn, ogn2017} method as shown in Figure~\ref{fig:overview}(e).
We only subdivide a local cell if it intersects with the input shape.
As the subdivision depth increases, the complexity of surface patch contained by each local cell decreases, leading to better approximation accuracy. 
In addition, since regions containing no shapes will not be further divided, we are able to accomplish a balanced trade-off between the computational complexity and reconstruction accuracy.
In all of our experiments, we use the octree-based subdivision for ground truth computation unless otherwise stated. Our experiments in Section~\ref{sec:exp_ablation} validate the benefits of performance from octree-based sampling.

\subsection{Surface Extraction}
\label{sec:surface_extract}

Once the network is learned, we are able to label each query point with our predictions. 
To extract the iso-surface, we first convert the inferred discrete labels back to the original 3PSDF representation. Points with labels 0, 1, and 2 are assigned with sdf values as -1, 1, and nan, respectively.
The reconstructed surface can then be extracted as zero-level surface.
Note that the iso-surface represented by 3PSDF can be directly extracted using the classic Marching Cubes (MC) algorithm.
The existence of null value would naturally prevent MC from extracting valid iso-surfaces at locations that contain no shapes. 
In the meantime, in the vicinity of target surface, the iso-surface extraction can be performed normally just as the conventional signed distance field.
After MC computation, we only need to remove all the nan vertices and faces generated by the null cubes.
The remaining vertices and faces serve as the meshing result.

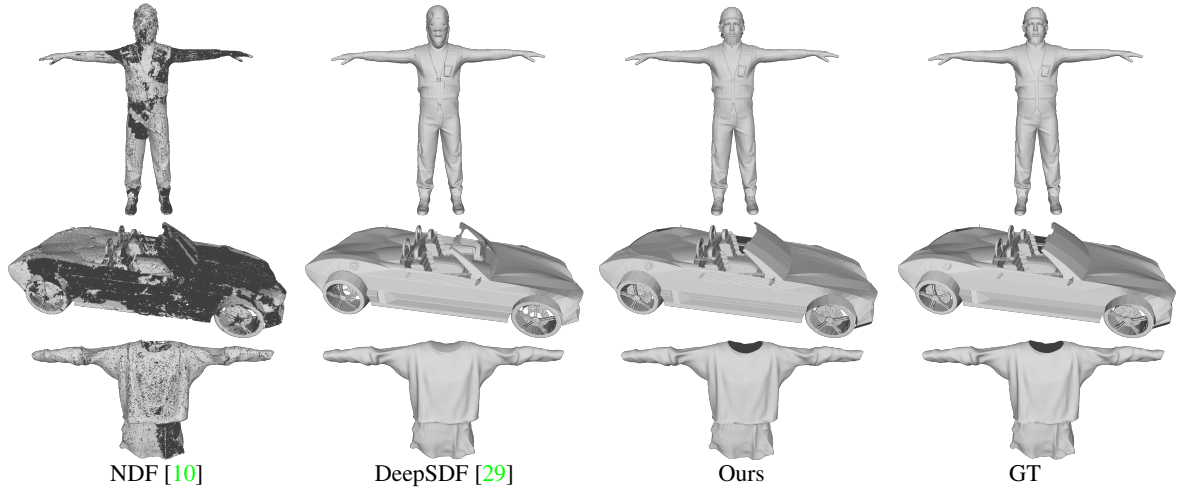
\begin{figure*}[!ht]
    \centering
    \vspace{-4mm}
    \begin{overpic}[width=0.9\linewidth]{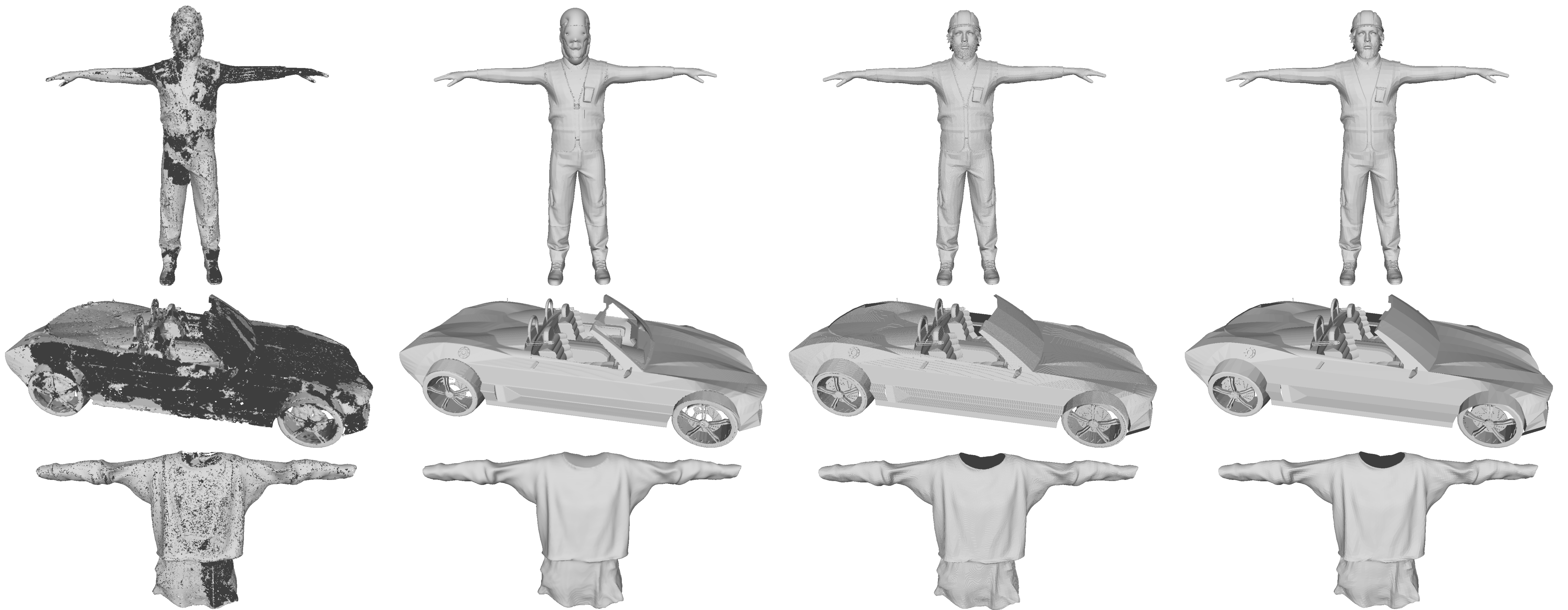}
    \put(40, -8) {\small NDF~\cite{chibane2020ndf}}
    \put(140,-8) {\small DeepSDF~\cite{park2019deepsdf}}
    \put(270, -8){\small Ours}
    \put(380, -8) {\small GT}
    \end{overpic}
    \vspace{1mm}
\caption{Visual comparisons of shape reconstruction result using different neural implicit representations.}
\label{fig:recon}
\vspace{-2mm}
\end{figure*}

\begin{table*}[!ht]
\begin{center}
\begin{tabular}{c||c||c|c|c|c|c||c||c}
    \hline
    \multirow{2}{*}{Metric} & \multirow{2}{*}{Method} & \multicolumn{5}{c||}{ShapeNet} & \multirow{2}{*}{MGN} & \multirow{2}{*}{Mixamo}\\
     \cline{3-7} & & car & plane & boat & lamp & chair & & \\
    \hline
    \hline
    \multirow{3}{*}{CD ($\times10^{-5}$) $\downarrow$} & NDF 
              & 0.63 & 0.25 & 0.33 & 0.34 & 0.45 & 0.08 & 0.52 \\
     & DeepSDF    & 2.71 & 0.58 & 0.61 & 1.99 & 0.91 & 0.09 & 1.82\\
     & Ours & \textbf{0.44} & \textbf{0.21} & \textbf{0.24} & \textbf{0.30} & \textbf{0.35} & \textbf{0.07} & \textbf{0.32}\\
    \hline
    \multirow{3}{*}{EMD ($\times10^2$) $\downarrow$} & NDF 
              & 2.39 & 2.46 & 2.11 & 2.05 & 1.47 & 0.33 & 2.81 \\
     & DeepSDF    & 4.23 & 2.56 & 2.29 & 2.78 & 1.66 & 0.62 & 4.56\\
     & Ours & \textbf{2.10} & \textbf{2.23} & \textbf{2.04} & \textbf{1.92} & \textbf{1.38} & \textbf{0.21} & \textbf{2.55} \\
    \midrule
\end{tabular}
\end{center}
\vspace{-7mm}
\caption{Quantitative comparisons of shape reconstruction using different neural implicit representations.}
\label{tab:shape_recon}
\end{table*}
\begin{figure*}[ht]
    \centering
    \begin{overpic}[width=0.9\linewidth]{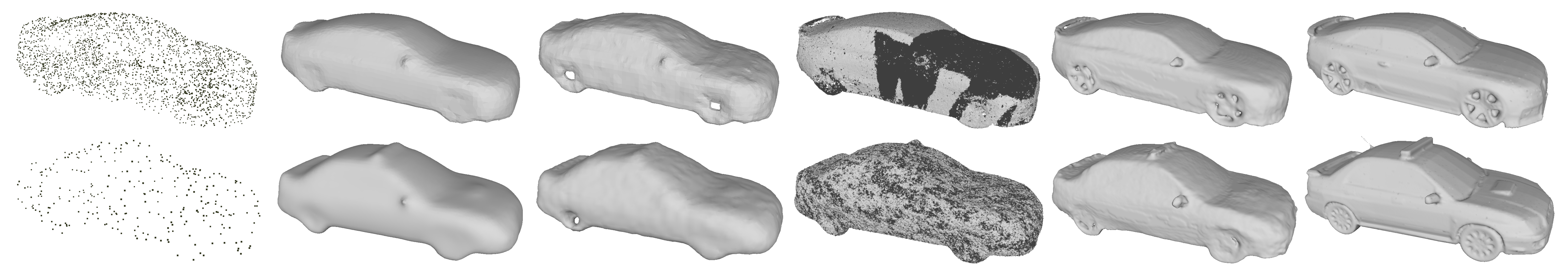}
    \put(30, -8) {\small Input}
    \put(100,-8) {\small OccNet~\cite{mescheder2019occupancy}}
    \put(180, -8){\small IF-Net~\cite{chibane20ifnet}}
    \put(250, -8) {\small NDF~\cite{chibane2020ndf}}
    \put(320, -8) {\small Ours}
    \put(400, -8) {\small GT}
    \end{overpic}
    \vspace{2mm}
    
\caption{Comparisons of point cloud completion trained on watertight shapes (with inner structure removed).}
\label{fig:point_closed}
\vspace{-3mm}
\end{figure*}

	\section{Experiments}
\label{sec:experiments}

\subsection{Experimental Setup}

\paragraph{Tasks and datasets.} 
We validate the proposed 3PSDF using three types of experiments.
First, we analyze the representation power of 3PSDF by examining how the 3PSDF can reconstruct complex 3D shapes from a learned latent embedding. This gives us an upper bound on the results we can achieve when conditioned on other inputs.
Second, we condition the learning of 3PSDF on sparse point cloud and test its performance by feeding 3D features.
Finally, we use image features as input and provide validation on the challenging task of singe-view reconstruction.
All experiments are compared with the SOTA methods for better verification.
The experiments are conducted on a wide range of 3D datasets including ShapeNet~\cite{chang2015shapenet}, MGN~\cite{bhatnagar2019mgn}, 3D-Front~\cite{fu20203d}, and Maximo~\cite{Mixamo}. 
The specific settings are detailed in the following experiments.

\paragraph{Implementation details.} 
For the task of reconstruction from point cloud, we use the same point encoder (IF-Net) and hyper-parameters with NDF~\cite{chibane2020ndf}.  
For single-view reconstruction, we use VGG16 \cite{simonyan2014vgg} with batch normalization as the image encoder. Similar to DISN \cite{xu2019disn}, we use both multi-scale local and global features to predict the 3PSDF value. 
We re-orient the normals of the ground-truth surfaces based on visibility \cite{chu2019repairing} to make them consistent. We refine the results by filling small holes and smoothing the surfaces. The ground-truth 3PSDF values are generated with resolution $128^3$ and the results are evaluated using resolution $256^3$.
We use octree-based importance sampling for all experiments.
To ease the learning of 3PSDF, we ensure the size of the minimum leaf octree cell to be consistent across different objects by using a unified bounding box for all samples.

\subsection{Shape Reconstruction}
\label{sec:exp_shape_recon}

\begin{figure*}[!htb]
    \centering
    \begin{overpic}[width=0.9\linewidth]{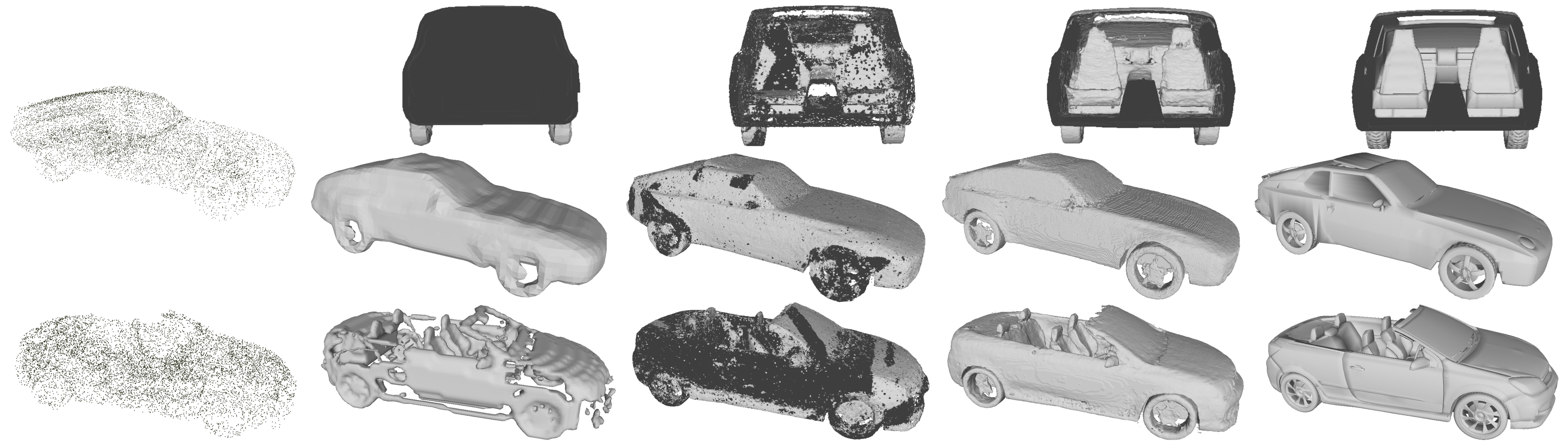}
    \put(30, -8) {\small Input}
    \put(120,-8) {\small SAL~\cite{Atzmon_2020_CVPR}}
    \put(210, -8){\small NDF~\cite{chibane2020ndf}}
    \put(300, -8) {\small Ours}
    \put(390, -8) {\small GT}
    \end{overpic}
\vspace{2mm}
\caption{Comparisons of point cloud completion trained on non-watertight shapes (with inner structure and open surface). The first row shows the inner structure of the reconstructed results in the second row.}
\label{fig:point_raw}
\end{figure*}

\begin{figure*}[h]
    \centering
    \begin{overpic}[width=0.9\linewidth]{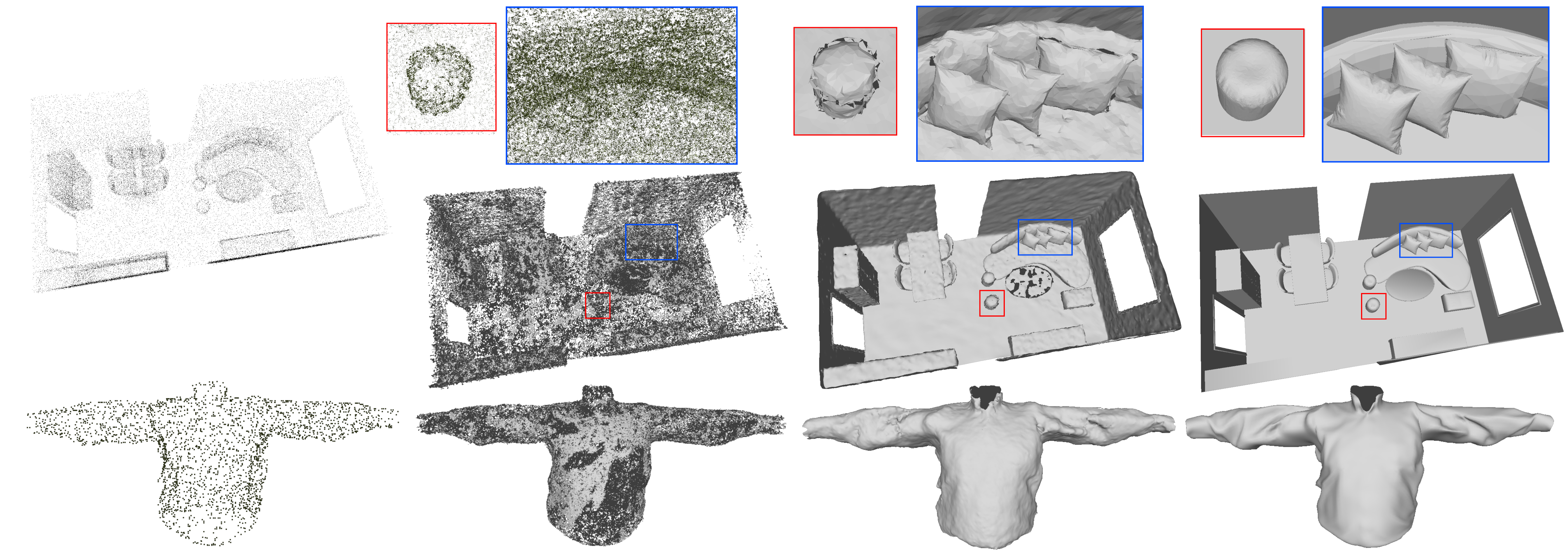}
    \put(50, -10) {\small Input}
    \put(160,-10) {\small NDF~\cite{chibane2020ndf}}
    \put(270, -10){\small Ours}
    \put(380, -10) {\small GT}
    \end{overpic}
    \vspace{2mm}
    
\caption{Comparisons of point cloud completion performance on MGN and 3D-Front. Chamfer-$L_2$ ($\times10^{-4}$) comparison: MGN: NDF - 0.035; ours - \textbf{0.033}; 3D-Front: NDF - 1.452; ours - \textbf{1.378}.}
\label{fig:point_mgn_scene}
\end{figure*}

To evaluate the capability of 3PSDF of modeling complex geometry, we perform the shape reconstruction experiment comparing with other SOTA neural implicit representations: DeepSDF~\cite{park2019deepsdf} and NDF~\cite{chibane2020ndf}.
Similar to the auto-encoding method in \cite{park2019deepsdf}, we embed each training sample with a 512 dimensional latent code and train neural networks to reconstruct the 3D shape from the embedding.
We perform evaluations on five representative categories of ShapeNet that contain the most intricate geometry, and two datasets with open surfaces: MGN~\cite{bhatnagar2019mgn} and Mixamo~\cite{Mixamo}.

Since we are only interested in reconstructing the training data, we do not use validation and test set for this experiment.
As DeepSDF cannot handle open surfaces, we generate its ground-truth SDF value using \cite{jacobson2013robust} which converts complex open surfaces into closed ones using winding number.
For training and evaluation, we use 10 as the depth for octree-based sampling for our method and the equivalent resolution of 1024 for DeepSDF.
To ensure similar density of sampling, we generate 1 million surface points for the NDF.
All the NDF results (including the following experiments) are generated using the post-processing scripts released by the authors to ensure fair comparison.
We show the visual comparisons in Figure~\ref{fig:teaser} and \ref{fig:recon} and the quantitative comparisons in Table~\ref{tab:shape_recon}.
While DeepSDF is able to reconstruct fine details, it cannot handle open surfaces like hair, clothing, and the windshield. 
NDF can deal with all topologies, but suffers from meshing problems -- lots of self-intersections and flipped faces are introduced.
Our method can faithfully reconstruct all the intricate geometries while achieving the best performance in quantitative comparisons.

\nothing{
	\note{
		\begin{itemize}
			\item Goal: Evaluate the upper bound of reconstruction accuracy of each baseline approach. We want to show 1) 3PSDF has much higher upper bound than NDF; 
			2) 3PSDF has comparable performance with traditional signed distance field for closed shapes; 3) 3PSDF performs much better than traditional signed distance field in open surface
			\item Setting: Given a 3D shape, we first convert it into baseline representation and then reconstruct it using field-to-mesh conversion.
			\item Dataset: ShapeNet for closed shape, MGN, Mixamo and 3D-Front for open surfaces
			\item Baselines: 1) NDF and 2) traditional signed distance field with marching cubes 
		\end{itemize}
	}
}

\subsection{Reconstruction from Point Cloud}
\label{sec:exp_point_cloud}

\begin{table}[!htb]
	\begin{minipage}[t]{.5\linewidth}
		\centering
		\begin{tabular}[t]{l||r||r}
			\hline
			& \multicolumn{2}{c}{Chamfer-$L_2$} \\
			\hline
			& \multicolumn{1}{c||}{3K} & \multicolumn{1}{c}{300} \\
			\hline
			\hline
			DMC     & 1.255         &  2.417 \\
			OccNet  & 0.938         & 1.009 \\
			IF-Net  & 0.326         &  1.147\\
			NDF     & 0.127         & 0.626 \\
			Ours    &\textbf{0.112} &  \textbf{0.595} \\
			\midrule
		\end{tabular}%
	\end{minipage}%
	\begin{minipage}[t]{.5\linewidth}
		\centering
		\begin{tabular}[t]{l||r||r}
			\hline
			& \multicolumn{2}{c}{Chamfer-$L_2$} \\
			\hline
			& \multicolumn{1}{c||}{10K} & \multicolumn{1}{c}{3K} \\
			\hline
			\hline
			SAL  & 6.39             & 7.39 \\
			NDF  & 0.074            & 0.275 \\
			Ours & \textbf{0.071}   & \textbf{0.258} \\
			\midrule
		\end{tabular}%
	\end{minipage} 
	\vspace{-3mm}
\caption{Left: results of point cloud completion for closed watertight cars from 3000 and 300 points. Right: results of point cloud completion for unprocessed cars from 10000 and 3000 points. Chamfer distance is  reported in $\times 10^{-4}$.}
\label{tab:point_car}
\end{table}

We further validate 3PSDF on the task of shape reconstruction from sparse point clouds.
Following NDF~\cite{chibane2020ndf}, we first evaluate 3PSDF on reconstructing closed surfaces, and then demonstrate that 3PSDF can represent complex surfaces with inner structures and open surfaces. 

\paragraph{Reconstruction of closed shapes.}
To compare with the SOTA methods: OccNet~\cite{mescheder2019occupancy}, IF-Net~\cite{chibane20ifnet}, and DMC~\cite{Liao2018CVPR}, we train on the ShapeNet car category pre-processed by  \cite{xu2019disn} with all open surfaces closed and inner structures removed.
We show the reconstruction results using 300 and 3000 points as input both qualitatively and quantitatively in Figure~\ref{fig:point_closed} and Table~\ref{tab:point_car} respectively.
Compared to the other methods, our approach can better reconstruct the sharp geometry details while outperforming all baselines in quantitative measurement.

\vspace{-4mm}
\paragraph{Reconstruction of complex surfaces.} 
To validate the ability of 3PSDF of handling raw, unprocessed data, we train 3PSDF to reconstruct complex shapes from sparse point clouds on three datasets: unprocessed cars from ShapeNet~\cite{chang2015shapenet}, garments with open surfaces from MGN~\cite{bhatnagar2019mgn}, and the living room scenes from 3D-Front~\cite{fu20203d}.
We use NDF~\cite{chibane2020ndf} and SAL~\cite{Atzmon_2020_CVPR} as the baselines for reconstructing unprocessed cars. 
Since SAL is built upon traditional SDF, we use the closed shapes as ground truth.
We provide visual comparisons of the reconstructed results in Figure~\ref{fig:point_raw} and \ref{fig:point_mgn_scene}.
SAL struggles to model the open surfaces, e.g. the windshield and the thin outer structure of car. 
NDF can generate dense point clouds close to the target surface. However, the output points are prone to be clustered (as shown in the closeups of Figure~\ref{fig:point_mgn_scene}) which prevent the BP algorithm from generating high-quality meshing results.
In contrast, 3PSDF is able to faithfully reconstruct the interior structures as well as the open surfaces. 
The quantitative comparisons in Table~\ref{tab:point_car} and Figure~\ref{fig:point_mgn_scene} also validates our advantage over the baselines.

\nothing{
\begin{itemize}
    \item Goal: We want to show 3PSDF is easier to learn comparing with NDF and other candidate approaches. Even for application like 3D reconstruction from sparse point cloud, which is the major application that NDF shows in their paper, our approach is still superior. 
    \item Setting: Given a sparse point cloud, the task aims to predict a complete mesh from the input.
    \item Dataset: MGN + ShapeNet + 3D-Front
    \item Baselines: IF-Net + NDF + SAL + L3PSDF w. SAL loss 
\end{itemize}
}

\subsection{Single-view 3D Reconstruction}
\label{sec:exp_mvr}

\nothing{
\begin{itemize}
    \item Goal: We want to show 3PSDF is easier to learn than the other baseline approach and is able to reconstruction shapes with complex topologies, i.e. containing both closed and open surfaces, in the challenging task like single/multi-view 3D reconstruction.
    \item Setting: Reconstruct complete 3D shape from given multi-view images
    \item Dataset: MGN + ShapeNet + 3D-Front
    \item Baselines: IF-Net + NDF + SAL 
\end{itemize}
}

In this experiment, we apply 3PSDF to single-view 3D reconstruction (SVR) tasks to further demonstrate its representational ability. We evaluate on the MGN dataset~\cite{bhatnagar2019mgn} and ShapeNet \cite{chang2015shapenet}. We use 
Chamfer-$L_2$ distance and F-score ($\tau=1\%$ volume diagonal length) as the evaluation metrics.

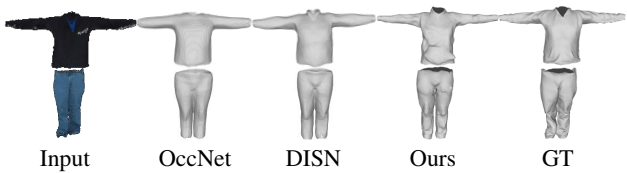
\begin{figure}[!htb]
    \centering
    \begin{overpic}[width=\linewidth]{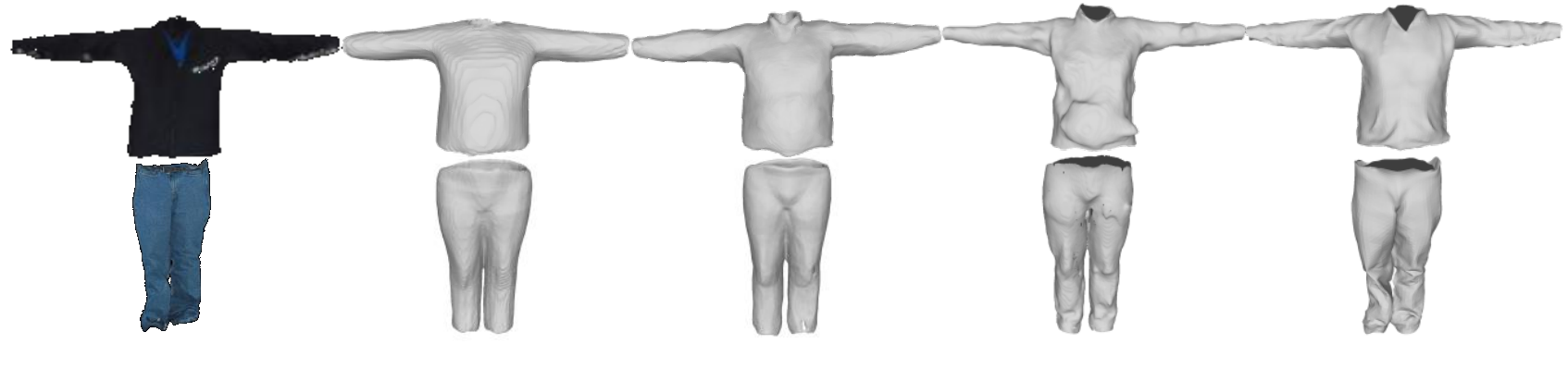}
    \put(15, -5) {\small Input}
    \put(60,-5) {\small OccNet}
    \put(108, -5){\small DISN}
    \put(155, -5) {\small Ours}
    \put(205, -5) {\small GT}
    \end{overpic}
    
\vspace{0mm}
\caption{Qualitative comparison on MGN dataset with state-of-the-art single-view reconstruction methods based on implicit functions. The quantitative evaluation results in terms of CD ($\times 10 ^{-3}$) and F-score ($\times 10^{-2}$ ) metrics on the testing set of MGN are: 1.03 and 69.8 (DISN); 1.01 and 71.0 (OccNet); \textbf{0.98} and \textbf{71.2} (Ours).}
\label{fig:svr_mgn}
\end{figure}

We compare against the representative SVR methods using implicit fields, including IMNet~\cite{imnet2019}, OccNet~\cite{mescheder2019occupancy} and DISN~\cite{xu2019disn}. We further implement an image-based NDF \cite{chibane2020ndf} estimator but find reasonable results cannot be generated by solely using image features. Since the models in these two datasets usually contain non-watertight surfaces which cannot be directly handled by the baseline methods, we first convert these models to watertight ones. Note that our representation is directly trained on the original shapes without this extremely time-consuming process. 

\vspace{-4mm}
\paragraph{Single-view reconstruction on MGN.} 
The models in the MGN dataset~\cite{bhatnagar2019mgn} are represented as open freeform surfaces with single sheets, which is challenging to the existing single-view reconstruction methods with implicit functions. We render an RGB image using the textured mesh for each garment model, and train a network conditioned on images to predict the shape representations. As shown in Figure~\ref{fig:svr_mgn}, our results capture the original open-surface structure as well as more high frequency geometric features such as the wrinkles. The 3PSDF representation also achieves the best quantitative results on the testing set.  

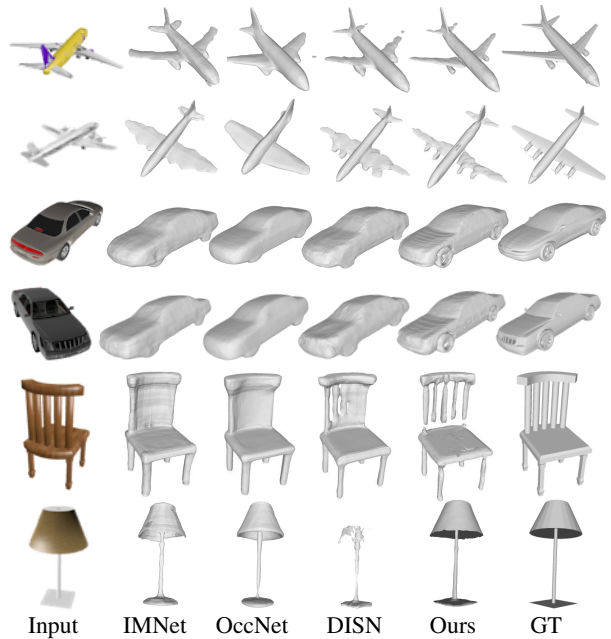
\begin{figure}[!htb]
    \centering
    \begin{overpic}[width=\linewidth]{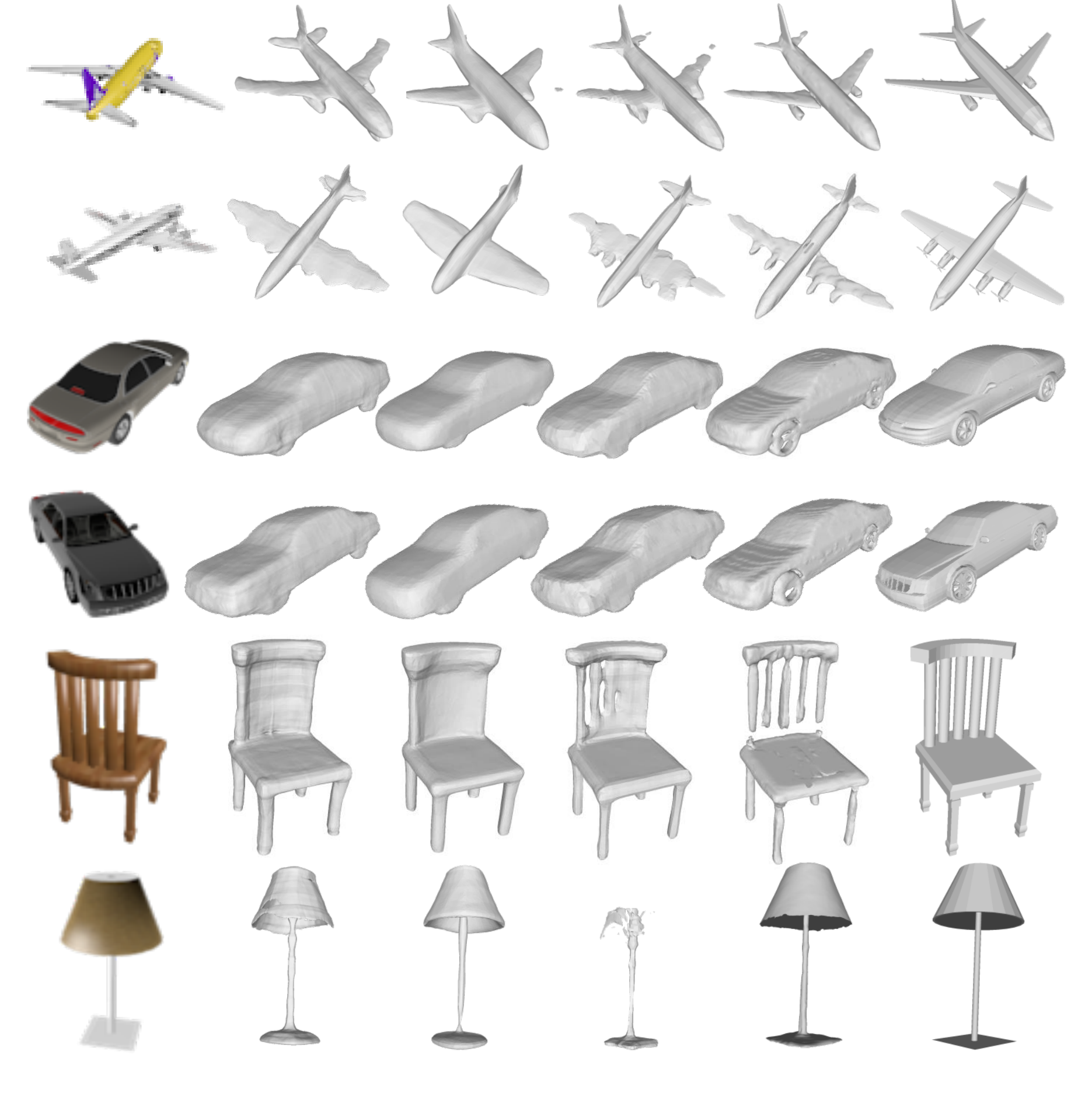}
    \put(13, 0) {\small Input}
    \put(49, 0) {\small IMNet}
    \put(84, 0){\small OccNet}
    \put(126, 0){\small DISN}

    \put(165, 0) {\small Ours}
    \put(203, 0) {\small GT}
    \end{overpic}
    
\vspace{-2mm}
\caption{Qualitative comparison results with SOTA single-view reconstruction methods based on implicit functions.}
\label{fig:svr}
\vspace{-4mm}
\end{figure}

\begin{table}[h]
\begin{center}
\begin{tabular}{c||c||c|c|c|c|c}
    \hline
     & \multirow{2}{*}{Method} & \multicolumn{5}{c}{ShapeNet}  \\
     \cline{3-7} & & car & plane & boat & lamp & chair  \\
    \hline
    \hline
    \multirow{3}{*}{CD $\downarrow$} 
    & IMNet     & 3.48 & 5.07  & 4.17 & 9.51  & 1.81   \\
     & OccNet    & 1.74 & 1.74 & 3.48  & 14.55  & 2.22  \\
     & DISN    & 1.23  & 1.71  & 4.84  & \textbf{6.11} & \textbf{1.54}   \\
     & Ours & \textbf{0.76}  & \textbf{1.66} &  \textbf{3.27} & 7.67  & 3.29   \\
    \hline
    \multirow{3}{*}{FS  $\uparrow$} 
    & IMNet   & 31.8  & 33.7 & 39.8 & 34.3  & 61.1  \\
     & OccNet    &54.4  & 59.7  & 44.9  &  \textbf{50.6} & 59.6  \\
     & DISN    & 65.8  & \textbf{77.2}  & 57.8 & 50.4 & \textbf{63.8}   \\
     & Ours & \textbf{77.0} & 72.8 & \textbf{66.6} & 49.3 & 58.5  \\
    \midrule
\end{tabular}
\end{center}
\vspace{-5mm}
\caption{Quantitative comparisons of single-view reconstruction. Chamfer-$L_2$ and F-score are reported in $\times 10^{-3}$ and $\times 10^{-2}$ respectively.}
\label{tab:svr}
\vspace{-3mm}
\end{table}


\paragraph{Single-view reconstruction on ShapeNet.}  We use a subset of ShapeNet \cite{chang2015shapenet} for evaluation, from which we choose 5 categories (airplane, car, lamp, chair, boat) resulting in 17803 shapes. We use the same image renderings (24 views per shape) and train/test split as Choy et al. \cite{choy20163d}.  
Figure~\ref{fig:svr} shows a set of qualitative comparisons. Despite being designed for handling open surfaces, 3PSDF is still a versatile representation for reconstructing various 3D shapes in the ShapeNet with either closed or open surfaces. We not only faithfully preserve the original structure of the target shape, but also captures more detailed geometries. Instead, the existing implicit functions always rely on watertight shapes, which substantially limits their representational ability and usually leads to over-smoothed geometries, lack of details, as well as inconsistent typologies. 
As shown in Table~\ref{tab:svr}, 3PSDF achieves state-of-the-art performance compared to the existing methods, where it has 5 metrics ranking the first and comparable results for the remaining metrics.



\subsection{Further Discussions}
\label{sec:exp_ablation}

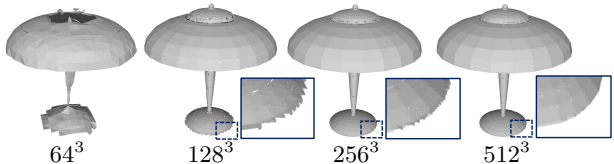
\begin{figure}[!htb]
    \centering
    \begin{overpic}[width=\linewidth]{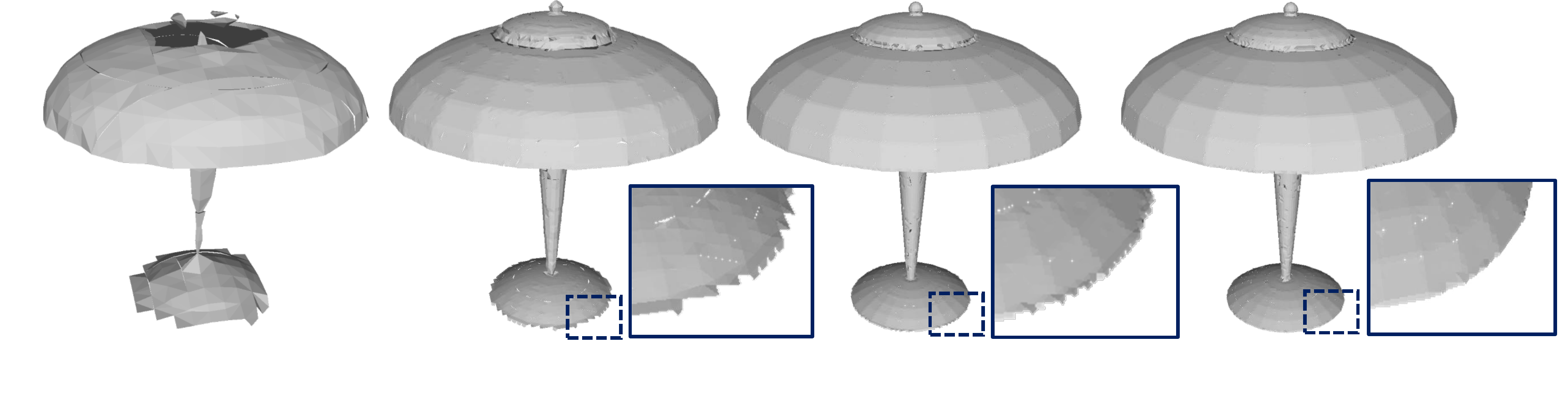}
    \put(23, 1) {\small $64^3$}
    \put(75,1) {\small $128^3$}
    \put(130, 1){\small $256^3$}
    \put(187,1) {\small $512^3$}
    \end{overpic}
    
\vspace{-3mm}
\caption{Reconstruction results of a shape using different resolutions. From the left to right, the CD ($\times 10^{-5}$) values for these shapes are: 14.49, 2.52, 2.21 and 2.12; the EMD ($\times 10^{2}$) values are: 3.42, 0.336, 0.267 and 0.227 .}
\label{fig:ablation_resoltuion}
\vspace{-6mm}
\end{figure}

\paragraph{Reconstruction accuracy/appearance with different resolutions.}
Since 3PSDF is continuously defined in the 3D space, it can represent a shape using arbitrary resolution. Figure~\ref{fig:ablation_resoltuion} gives a coarse-to-fine shape approximation results, where we discretize the volumetric space and use different grid resolutions to represent a 3D shape. The experimental results show that the approximation quality of 3PSDF increases as the resolution grows, leading to smoother shape boundaries and higher reconstruction accuracies.

\paragraph{Timing cost for field-to-mesh conversion.}
We quantitatively evaluate the timing cost for field-to-mesh conversion in different output sampling densities. For the octree depth of 6 ($64^3$), 7 ($128^3$), 8 ($256^3$) and 9 ($512^3$), the average field-to-mesh conversion times of 3PSDF for a single shape are 0.006s, 0.11s, 0.54s, and 3.72s respectively. In contrast, the conversion times for NDF~\cite{chibane2020ndf} given the comparable number of sampling points are: 2.1s, 15mins, 3hrs, 34hrs, using the provided post-processing setting (radius=0.005) by NDF. The experiments are conducted on a machine with a 48-Core AMD EPYC CPU and 64GB memory.

\nothing{
\begin{itemize}
    \item Goal: We want to show our method can achieve much faster field-to-mesh conversion compared to the approaches based on unsigned distance field, e.g. NDF.
    \item Setting: Given a field representation of an open surface, compute the time of converting it into corresponding mesh.
    \item Dataset: 3D-Front + Mixamo + MGN
    \item Baselines: NDF 
\end{itemize}
}

\begin{table}[htbp]
  \centering
    \begin{tabular}{lrrr}
    \hline
          & Random  & Uniform & Octree \\
    \hline
    CD $(\times10^{-4})$ $\downarrow$    & 7.43  & 2.16  & \textbf{1.08} \\
    EMD $(\times10^3)$ $\downarrow$   & 3.28  & 1.55  & \textbf{1.12} \\
    \midrule
    \end{tabular}
  \vspace{-3mm}
  \caption{Reconstruction accuracy using different sampling strategies.}
   \vspace{-6mm}
  \label{tab:sampling}
\end{table}%

\paragraph{Different sampling strategies.}
We further study the impact of different sampling strategies on the performance of 3PSDF; we evaluate on the task of shape reconstruction from point cloud on the unprocessed car data.
Three strategies are used to generate sampling points: 1) randomly draw samples in the space; 2) uniform sampling which generates adjacent points in equal distance; 3) octree-based sampling that uses the corner points of leaf octree cells as training samples. We use around 18 million sampling points for all strategies.
Table~\ref{tab:sampling} shows that the octree-based sampling yields the best result.
Compared to the other methods, octree-based sampling is able to densely sample points with inside/outside labels, generating a more balanced training set containing all the 3 labels. We use octree-based sampling for all of our experiments unless otherwise stated.

\vspace{-4mm}
\paragraph{Limitation.} 3PSDF has difficulty in reconstructing  multi-layer surfaces that are very close to each other, especially when the resolution is low. 
This is because 3PSDF requires denser sampling rate compared to SDF in order to insert a null layer in between to prevent artifact surface.
Besides, given the enhanced representational ability of 3PSDF, it requires more informative features to learn and longer time to train; for example, the network converges much faster and achieves better geometry given point clouds as input, compared to single images.

	\section{Conclusions and Discussions}
\label{sec:conclusion}
We introduce 3PSDF, a learnable implicit distance function to represent 3D shapes with arbitrary topologies. Different from the widely used implicit representations like SDF that can only encode watertight shapes, 3PSDF can faithfully represent various shapes with both open and close surfaces. The key insight of the 3PSDF is the introduction of the NULL sign to additionally indicate the inexistence of surface. We further formulate a classification-based learning paradigm to effectively learn this representation. As a result, the representational power of the distance function is significantly enhanced. Extensive evaluations demonstrate that 3PSDF is a versatile implicit representation that accommodates various 3D reconstruction tasks.

\vspace{-4mm}
\paragraph{Future work.} We have shown in the supplemental that 3PSDF can be learned via an alternative method that combines binary classification and regression. 
Compared to 3-way classification, such a method has the potential to generate smoother surface with fewer sampling points at training time. However, it is not as robust as the 3-way counterpart as it requires the results of the two branches align well in order to prevent holes and artifacts.
It would be an interesting future avenue to investigate how to resolve this issue.


	
	{\small
		\bibliographystyle{ieee_fullname}
		\bibliography{paper}
	}
	
	\ifthenelse{\equal{\forArxiv}{1}}{
    	\clearpage
    	\appendix
    	In this supplemental material, we discuss an alternative framework of learning 3PSDF (Section~\ref{sec:supl_regress}), use 3PSDF to model functions or manifolds (Section~\ref{sec:supl_function}), provide additional implementation details (Section~\ref{sec:supl_imp_details}), network structure for each experiment (Section~\ref{sec:supl_network}), comparison between our proposed 3PSDF and TSDF (Section~\ref{sec:supl_tsdf}), and more results (Section~\ref{sec:supl_more_results}).
	
    	\section{Alternative Learning Framework}
\label{sec:supl_regress}

In addition to 3-way classification, 3PSDF can be learned using an alternative framework that combines binary classification and regression.
Specifically, the binary classification branch learns to classify the space into nan and non-nan regions, where the non-nan region forms a valid narrow band for extracting surface as demonstrated in Figure 2(b) as shown in the main paper.
The regression branch strives to regress a continuous SDF in the narrow-band region as generated by the classification branch. 
Formally, we formulate this alternative framework as follows:

\begin{equation}
  \Phi_C(\mathbf{p}, \mathbf{x}): \mathbb{R}^3 \times \mathcal{X} \mapsto [0, 1],
\label{eqn:bin_cls}
\end{equation}

\vspace{-4mm}
\begin{equation}
\Psi_R(\mathbf{p}, \mathbf{x}) = SDF(\mathbf{p}).
\label{eqn:regress}
\end{equation}

In particular, the classification branch $\Phi_C$ consumes a 3D query point $\mathbf{p}$ and its corresponding observation $\mathbf{x}$ and predicts the probability of the query point locating in the non-nan region; the regression branch $\Phi_R$ directly infers the signed distance of $\mathbf{p}$ as defined in Equation (3) in the main paper.

\begin{figure}[!ht]
    \centering
    \includegraphics[width=0.9\linewidth]{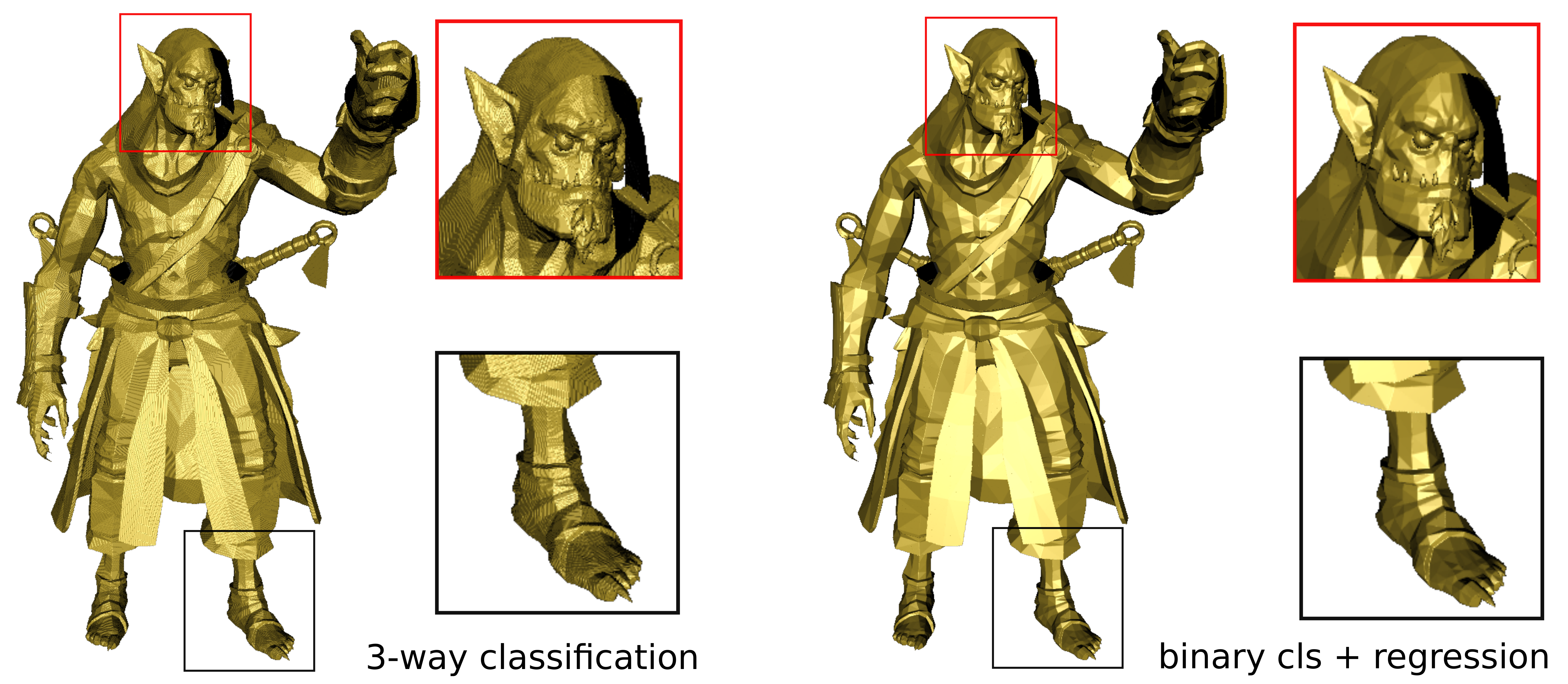}
\caption{Comparisons of two ways of learning 3PSDF. Quantitative comparisons of shape reconstruction, Mixamo: 0.32:0.31(CD); 0.944:0.950(F-score); MGN: 0.07:0.07(CD); 0.991:0.993(F-score). Note all numbers are reported in format of (3-way cls. : bin. cls.+reg.).}
\label{fig:rebut_regress}
\end{figure}

\paragraph{Surface extraction.} 
The framework based on binary classification and regression requires training of two branches, which can be implemented either using two heads of a backbone network or two independent networks.
Once the networks are trained, the sampling points that are classified as nan points by the classification branch are assigned with nan value.
The rest points are assigned with continuous SDF distance using the predictions of the regression branch.
The resulting 3PSDF field can be directly converted into mesh using the Marching Cubes (MC) algorithm with the iso-value set to 0. 
Same as 3-way classification, after MC computation, we only need to remove all the nan vertices and faces generated by the null cubes. The remaining vertices and faces serve as the meshing result.


\begin{figure*}[ht]
    \centering
    \begin{overpic}[width=\linewidth]{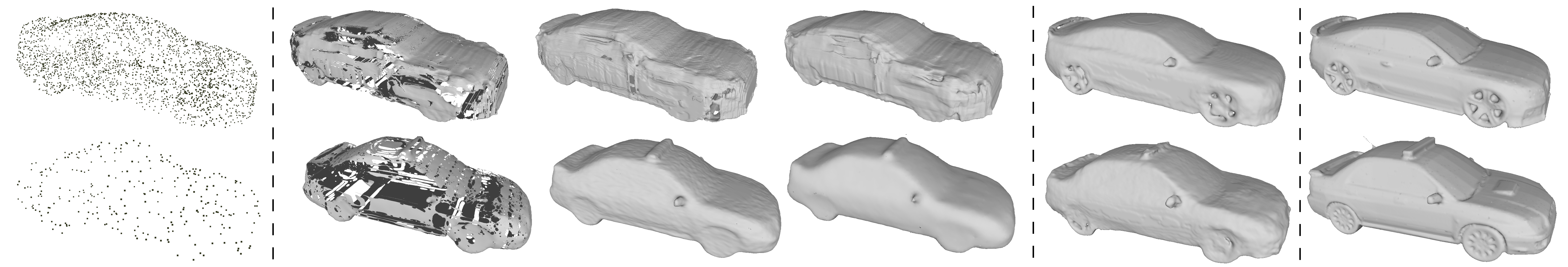}
    \put(35, -8) {\small Input}
    \put(103,-8) {\small Bin. cls.+Reg.}
    \put(195, -8){\small Bin. cls.}
    \put(270, -8) {\small Regression}
    \put(355, -8) {\small 3-way cls.}
    \put(450, -8) {\small GT}
    \end{overpic}
    \vspace{0.2mm}
\caption{Comparisons of point cloud completion trained on watertight shapes by using two candidate learning frameworks of 3PSDF: binary classification (bin. cls.)+regression (reg.) and 3-way classification (cls.). For the results of BR, we also show the results generated from the two branches.}
\label{fig:supp_reg_vs_cls}
\end{figure*}
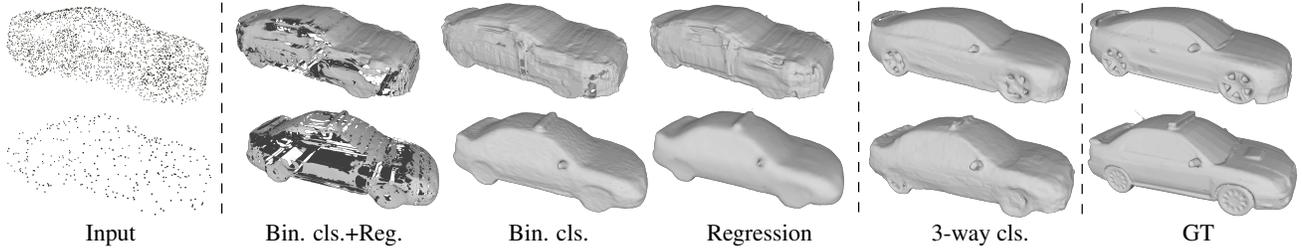

\subsection{Comparisons with 3-way Classification}

We provide in-depth comparisons between the two candidate learning frameworks: binary classification + regression (BR) v.s. 3-way classification (3C) in this section.
Specifically, we evaluate both methods in the task of shape reconstruction and point cloud completion.

\paragraph{Shape reconstruction.} 
We use the same experiment settings with that of the main paper for evaluating the two candidate frameworks.
Both methods are validated using two datasets that contain non-watertight open surfaces: MGN~\cite{bhatnagar2019mgn} and Mixamo~\cite{Mixamo}.

We show both the qualitative and quantitative comparisons in Figure~\ref{fig:rebut_regress}.
While the two methods are trained using the same data, the BR framework can generate smoother reconstruction compared to that of 3C method, thanks to its continuous SDF output. 
This is also reflected in the quantitative measurements, where BR can achieve comparable or even better results.

\begin{figure}[!htb]
    \centering
    \includegraphics[width=\linewidth]{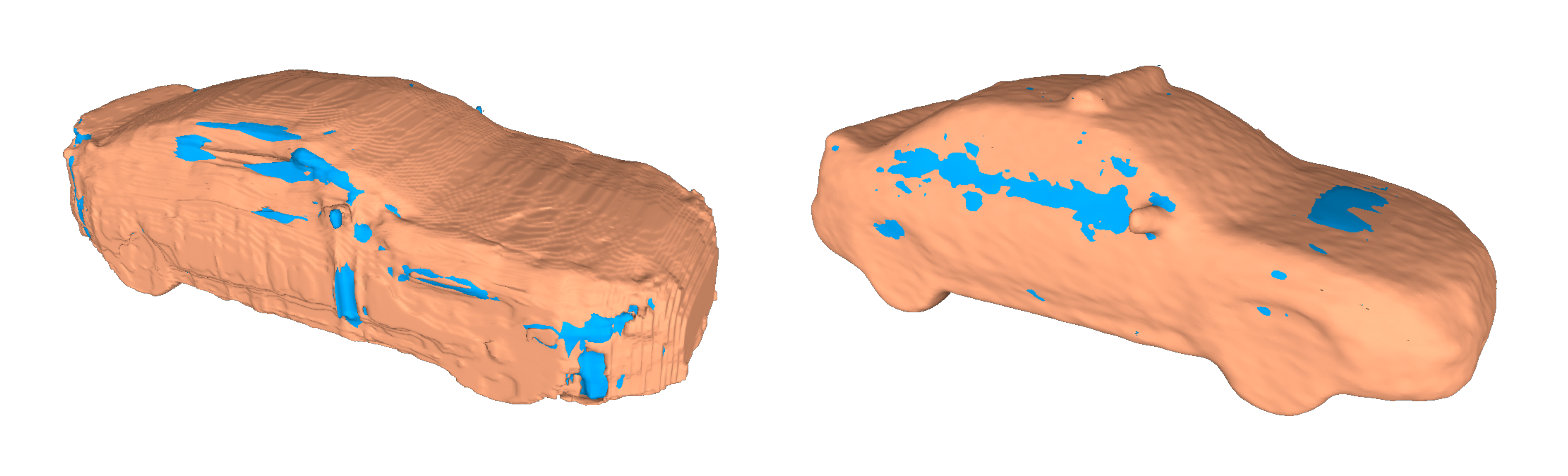}
\caption{We overlay the reconstruction results of the classification and regression branches under the BR framework as shown in Figure~\ref{fig:supp_reg_vs_cls}. The classification results are highlighted in orange while the regression results are marked with blue. The misalignment of the two branches' results leads to the incomplete reconstruction in Figure~\ref{fig:supp_reg_vs_cls}.}
\label{fig:supp_reg_overlay}
\end{figure}

\begin{table}[!htb]
	\begin{minipage}[t]{.5\linewidth}
		\centering
		\begin{tabular}[t]{l||r||r}
			\hline
			& \multicolumn{2}{c}{Chamfer-$L_2$} \\
			\hline
			& \multicolumn{1}{c||}{3K} & \multicolumn{1}{c}{300} \\
			\hline
			\hline
			BR     & 0.312         &  1.025 \\
			3C    &\textbf{0.112} &  \textbf{0.595} \\
			\midrule
		\end{tabular}%
	\end{minipage}%
	\begin{minipage}[t]{.5\linewidth}
		\centering
		\begin{tabular}[t]{l||r||r}
			\hline
			& \multicolumn{2}{c}{Chamfer-$L_2$} \\
			\hline
			& \multicolumn{1}{c||}{10K} & \multicolumn{1}{c}{3K} \\
			\hline
			\hline
			BR  & 0.095             & 0.314 \\
			3C & \textbf{0.071}   & \textbf{0.258} \\
			\midrule
		\end{tabular}%
	\end{minipage} 
	\vspace{-3mm}
\caption{Left: results of point cloud completion for closed watertight cars from 3000 and 300 points. Right: results of point cloud completion for unprocessed cars from 10000 and 3000 points. Chamfer distance is  reported in $\times 10^{-4}$.}
\label{tab:supp_reg_vs_3way}
\end{table}

\paragraph{Reconstruction from point cloud.} 
We also validate the performance of both candidate frameworks in the task of surface reconstruction from sparse point cloud. 
Specifically, we evaluate their performance on reconstructing both closed and open surface with the same setting as that of the main paper.
We show in Figure~\ref{fig:supp_reg_vs_cls} that in the BR framework, though both the classification and regression branches can generate reasonable reconstructions, the final merged results still exhibit incompleteness. 
We further demonstrate the cause of incomplete reconstructions in the overlaid visualization of the two branches (Figure~\ref{fig:supp_reg_overlay}). 
Since the results of the two branches are not perfectly aligned due to the different natures of their tasks, the classification branch would mistakenly remove part of the regressed surfaces generated by the regression branch.
This could render holes and discontinuity in the results of the BR method.
In comparisons, the 3C method does not suffer from such a problem as it only requires a single branch to generate the final reconstruction.
This is also reflected in the quantitative measurements in Table~\ref{tab:supp_reg_vs_3way}.

\paragraph{Discussion.}
We have evaluated the performance of both candidate frameworks in two different tasks. 
In the applications where the binary classification and regression branches are well aligned, e.g. the shape reconstruction task, the BR method can lead to higher-quality results with smoother surface compared to the 3C approach.
However, for more challenging scenarios, e.g. point cloud completion, where the two branches of BR framework may produce slightly deviated reconstructions, the final reconstruction may be incomplete despite that the two branches have obtained faithful reconstructions.
In contrast, the 3C framework is robust over all kinds of task without the need of worrying about the misalignment issue.
It would be an interesting future avenue to investigate how to resolve the misalignment problem of the BR method while enjoying its smooth nature.

    	\section{Modeling Functions and Manifolds using 3PSDF}
\label{sec:supl_function}

Following NDF, we train 3PSDF on 1 million points sampled from 1000 functions, which are either linear, parabola
or sinusoids. Figure~\ref{fig:rebut_function} shows the fitting results of 3PSDF to a variety of functions and manifolds. 
In Figure~\ref{fig:rebut_function}, red dots are points labeled as “inside” while cyan ones as “outside”.
“Nan” points are omitted for clear demonstration.
As shown in the results, 3PSDF can faithfully model various functions and manifolds, which further validate that it is a versatile representation.

\begin{figure}[!htb]
    \centering
    \vspace{-3mm}
    \includegraphics[width=\linewidth]{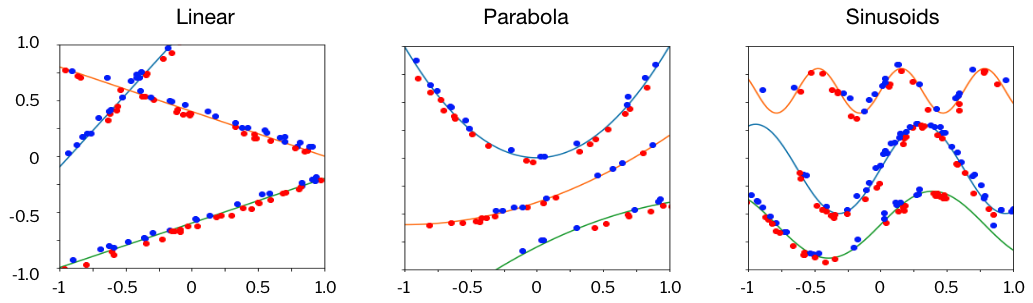}
\vspace{-6mm}
\caption{Function and manifold fitting using 3PSDF.}
\label{fig:rebut_function}
\end{figure}

        \section{More Implementation Details}
\label{sec:supl_imp_details}

\subsection{Reconstruction from Sparse Point Cloud}
\label{sec:suppl_point_complete}

We use octree-based sampling to generate the ground-truth data for our approach.
The sampling points are the corner points of the leaf cells generated by octree decomposition.
In particular, we use depth of 6 for generating training data on pre-processed ShapeNet car category.
For raw, unprocessed ShapeNet car, MGN, and 3D-Front, we use depth of 8, 7, and 9 respectively for training data generation.
We train separate models for different numbers of input points.
All models are trained using the same set of hyperparameters.
For all experiments, we use the Adam optimizer with parameters $lr = 1e^{-4}$, $betas = (0.9, 0.999)$, $eps = 1e^{-8}$, $weight\_decay = 0$.

For MGN dataset, we split the data into train and test set with 9:1 ratio.
For 3D-Front dataset, we extract 100 living rooms, 10 of which is used for testing and the rest is used for training.
For NDF, we generate 1 million points for all experiments except the scene reconstruction task where we generate a more dense point containing 3 million points.
The meshing results of NDF are obtained by running the script (including Ball Pivoting algorithm (BPA) and post-processing operations) provided by the authors in MeshLab.
All the results are reported using the test data. 
For the ShapeNet car dataset, we use the common train and test split by \cite{xu2019disn}.

\subsection{Single-view Reconstruction on MGN}
We evaluate and compare the representation capability of 3PSDF, DISN \cite{xu2019disn} and OccNet \cite{mescheder2019occupancy} on MGN dataset \cite{bhatnagar2019mgn} for single-view 3D reconstruction. Each garment model in MGN dataset is rendered into an $256\times 256$ RGB image from a front-view textured mesh. All the meshes and images are aligned with the same camera settings and normalized.

For 3PSDF, open surface models in MGN dataset are directly sampled with Octree-based subdivision at a resolution of $128^3$, resulting in a mean sampling points of 300k across all models. The training batch size 
is set to 8 and the number of sampling points is 10k per sample. We use Adam optimizer with initial learning rate of 3e-4 and exponentially decayed to 0.99 at every 10k steps. For DISN and OccNet, models in MGN dataset are first converted to watertight form and then sampled with the default strategies used in the original papers. Each watertight model is sampled with 300k points, equivalent to that in 3PSDF. All the other training hyperparameters are set to default values.

MGN dataset is split into training and testing datasets with 9:1 ratio, and all 3 networks are evaluated at 20k epoches.

\subsection{Single-view Reconstruction on ShapeNet}

We use 17803 shapes from 5 categories of ShapeNet \cite{chang2015shapenet} for evaluation, including Airplane, Car, Lamp, Chair and Boat. We use the same image renderings (24 views per shape) and train/test split as Choy et al. \cite{choy20163d}. 

We directly use the raw mesh of ShapeNet to generate the ground truth to train 3PSDF, while the competitive methods are trained using pre-processed watertight meshes. The ground truth 3PSDF values are sampled with resolution $128^3$ and the results are evaluated using resolution $256^3$. The images are all scaled to the resolution of 224 $\times$ 224. We first train the network for 30 epochs with learning rate 1e-4, and then finetune the network for 80 epochs using learning rate 5e-5. The batch size is set to 8 and the number of sampled points is 20k for one shape in each iteration during training. 
The reconstruction results are post-processed with simple hole filling and smoothing.

        \section{Network Structure}
\label{sec:supl_network}

\subsection{Network Architecture for Shape Reconstruction}

Figure~\ref{fig:supp_shape_recon_network} shows the detailed network structure for the experiment of shape reconstruction.
In particular, the network follows the design of the auto-decoder~\cite{park2019deepsdf} which does not requires an encoder for learning the shape priors of training data.
The input to the decoder contains: 1) a 512-dimensional per-object latent code, that is learned during training, and 2) a point feature obtained after applying point feature extractor to the 3D coordinate of the query point.

\begin{figure}[!htb]
    \centering
    \includegraphics[width=\linewidth]{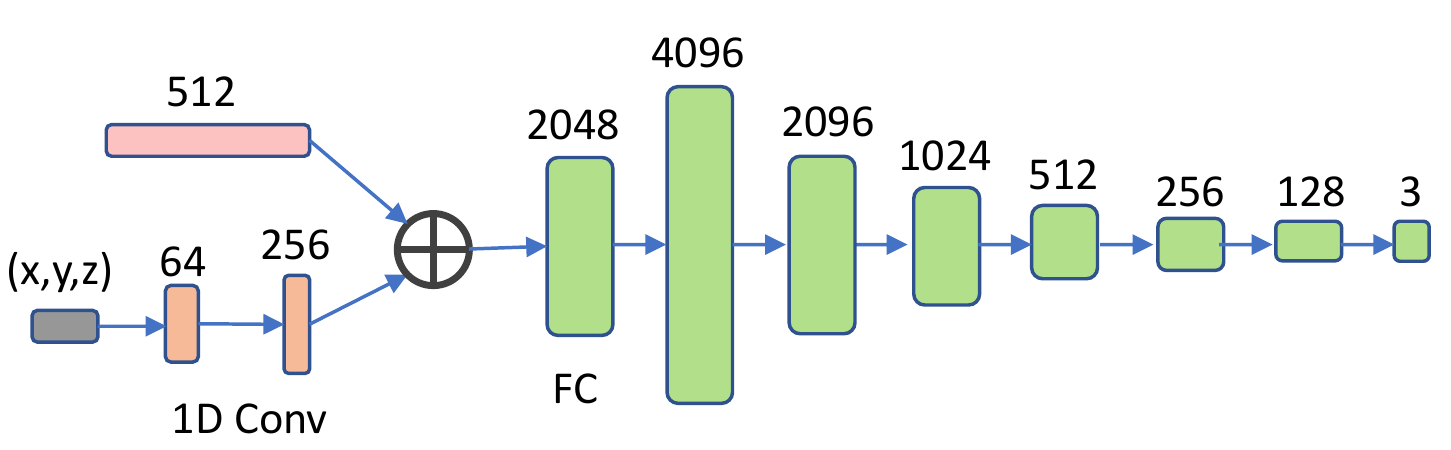}
\caption{Network structure for shape reconstruction.}
\label{fig:supp_shape_recon_network}
\end{figure}

The point feature extractor is implemented using 1D convolutional operator.
The concatenation of the latent code and the point feature is then fed into the decoder which consists of multiple fully connected layers.
The output layer of the decoder predicts the per-class probability for the 3 categories defined by 3PSDF.

\subsection{Network Architecture for Reconstruction from Point Cloud}

\begin{figure}[!htb]
    \centering
    \includegraphics[width=\linewidth]{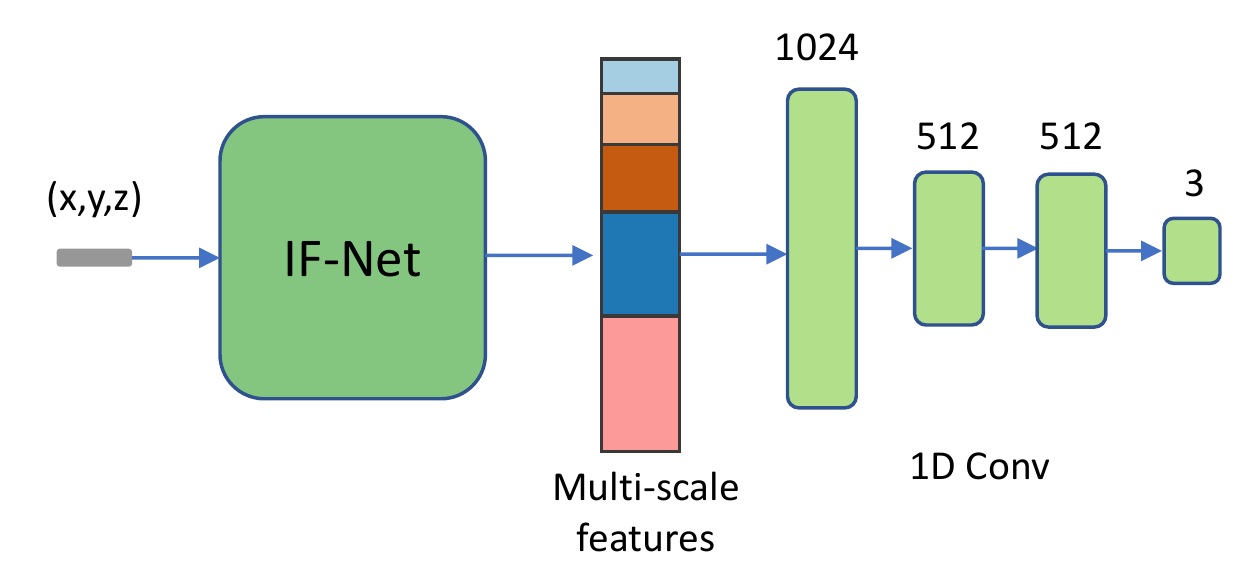}
\caption{Network structure for reconstruction from point cloud.}
\label{fig:supp_point_complete_network}
\end{figure}

We show the detailed network structure for reconstruction from point cloud in Figure~\ref{fig:supp_point_complete_network}.
To ensure fair comparison, we use the identical network with NDF~\cite{chibane2020ndf}, which is based on IF-Net~\cite{chibane20ifnet}, for extracting the features from the input point cloud.
The extracted multi-scale point features are then fed into the decoder.
The decoder is implemented using four 1D convolution layers, where the last layer predicts the per-class probability for 3PSDF.

\begin{figure}[!htb]
    \centering
    \includegraphics[width=\linewidth]{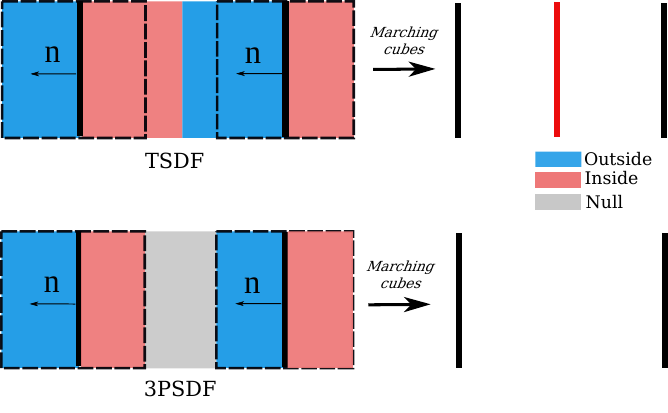}
\caption{Comparison between TSDF and the proposed 3PSDF. For reconstructing two adjacent single layers of mesh, TSDF would introduce artifacts (the red layer show on the right of first row) to the reconstruction result.}
\label{fig:supp_tsdf_compare}
\end{figure}

\begin{figure*}[t]
    \centering
    \begin{overpic}[width=\linewidth]{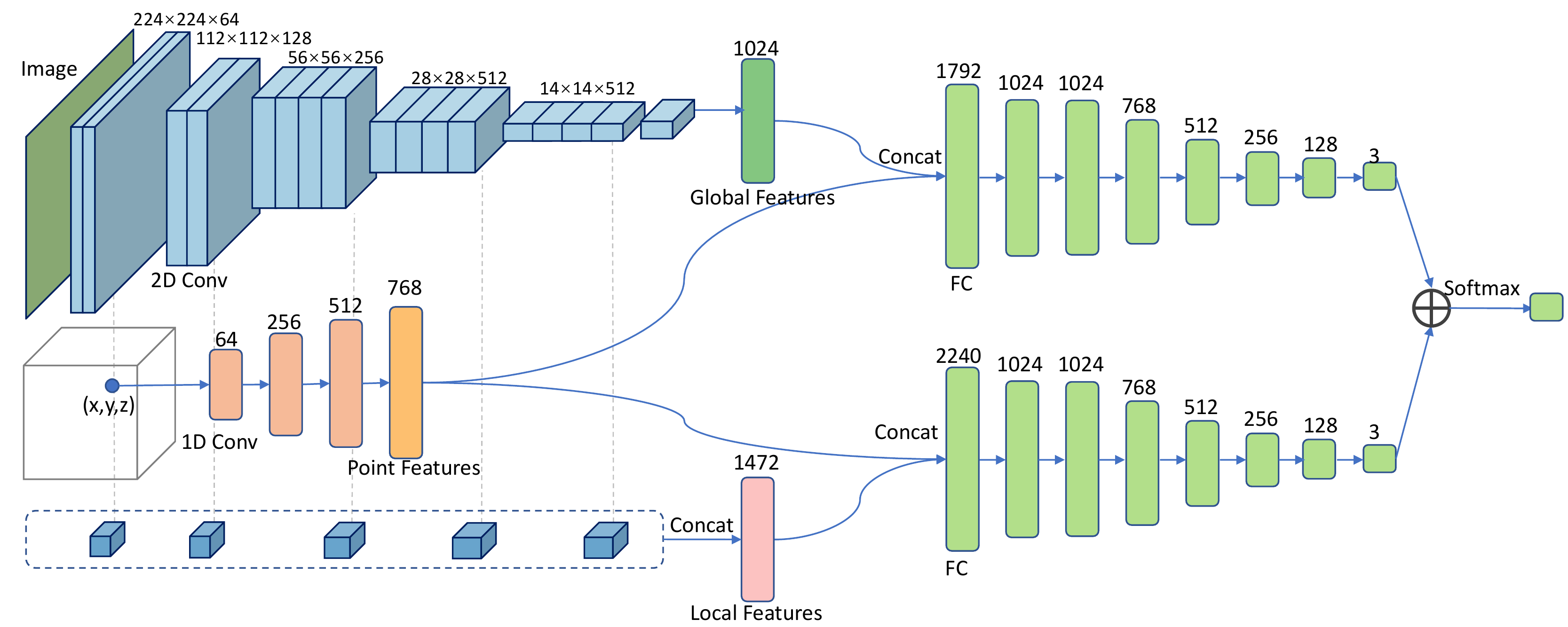}

    \end{overpic}
    \caption{Detailed network architecture for single view reconstruction.}
\label{fig:svr_network_architecture}
\vspace{-4mm}
\end{figure*}

\subsection{Network Architecture for SVR}

Figure~\ref{fig:svr_network_architecture} shows the detailed network architecture for 3D reconstruction based on single-view images. The network takes a set of sampled 3D points and a single view image as input. We use several 1D convolution layers to obtain the point features and a VGG-16 (with batch normalization) architecture to encode the input image. We adopt a two-stream network architecture, where the point features are concatenated with global and local image features respectively, and then fed into two branches to predict the 3PSDF. 

The global image features are obtained from an average pooling and a fully connected layer at the end of the image encoder. For the local features, we project the input 3D points to the image plane and retrieve the features on each feature map using the projected coordinates. The retrieved features on each feature map are concatenated together to obtain a local image feature vector. 

The decoder has two streams with the same structure, each of which consists of a set of fully connected layers to predict the 3PSDF separately. The outputs from the two branches are summed up and passed through a Softmax layer to obtain the final prediction.

\section{Comparison with TSDF}
\label{sec:supl_tsdf}

Truncated Signed Distance Field (TSDF) is widely used in obtaining reconstruction results from the volumetric range data, e.g. the RGBD stream from depth sensors. 
One mainstream application of TSDF is large-scale tracking and mapping in reconstructing 3D scenes.
As one may have seen open surfaces, e.g. the walls in the reconstructed 3D environment, can be reconstructed using TSDF, we provide detailed comparisons here stating the difference between TSDF and 3PSDF regarding the ability of modeling surfaces with arbitrary topologies.

The motivation of introducing TSDF is to set a lower bound of reconstruction error during the fusion of different SDFs converted from the depth maps.
In particular, in real-world scanning, the raw data obtained from the depth sensor is highly likely to be contaminated by the noises.
In practice, the depth maps are converted into SDFs in order to fuse the per-frame observation into a more complete reconstruction in the canonical space.
However, the most widely adopted way of fusing the SDFs is based on weighted summation, where the errors brought by each SDF would be accumulated and affecting the previously fused results.
TSDF alleviates this issue by clipping the minimum and maximum signed distance value and hence prevents the summed TSDFs from deviating too much from the ground-truth value.

After analyzing the motivation of TSDF, we can better understand the difference between TSDF and our proposed 3PSDF.
(1) Unlike 3PSDF, TSDF remains a binary-sided signed distance function which only has positive and negative signs.
This could render TSDF failed to represent open surfaces without introducing artifacts in many cases.
As shown in Figure~\ref{fig:supp_tsdf_compare} upper row, for two adjacent surfaces with consistent normals, the positive and negative signs would intersect with each other in the middle region where the SDFs are truncated to maximum and minimum respectively. 
This leads to an additional surface/artifact (the red boundary on the right) if meshing such a field using the Marching Cubes algorithm.
In contrast, 3PSDF can achieve artifact-free reconstruction by inserting a NULL layer in between to prevent the formation of the additional decision boundary.
(2) The way that TSDF models open surfaces is completely different from that of 3PSDF.
In particular, TSDF generates open surfaces by space clipping, where only the field within a bounded volume is converted into mesh.
In comparison, 3PSDF is able to model open surfaces by directly meshing the entire 3D space without requiring a clipping bounding volume.

        \section{More Results}
\label{sec:supl_more_results}

\paragraph{Reconstruction of closed surfaces from sparse point cloud.} 
We provide more qualitative comparison results with the state-of-the-art approaches on the task of shape reconstruction from sparse point cloud.
In Figure~\ref{fig:supp_point_complete_car_disn}, we show the reconstruction result using the models trained on preprocessed ShapeNet car data (watertight mesh with inner structure removed) provided by \cite{xu2019disn}.

\paragraph{Reconstruction of complex surfaces from sparse point cloud.}
In Figure~\ref{fig:supp_point_complete_car_raw} we provide more qualitative comparisons of shape reconstruction results of complex surfaces that contain both closed and open surfaces.
All the candidate approaches, including ours, are trained on on raw, unprocessed ShapeNet car data, which contain inner structures and open surfaces.
As seen in the highlighted regions within the red rectangles, our approach is able to generate shapes with consistent normals even when the ground truth data may contain flipped face patches.

\begin{figure*}[!htb]
    \centering
    \begin{overpic}[width=\linewidth]{figs/raster/suppl_car_disn.pdf}
    \vspace{8mm}
    \put(35, 7) {\small Input}
    \put(105, 7) {\small OccNet~\cite{mescheder2019occupancy}}
    \put(185, 7){\small IF-Net~\cite{chibane20ifnet}}
    \put(270, 7){\small NDF~\cite{chibane2020ndf}}
    \put(360, 7) {\small Ours}
    \put(450, 7) {\small GT}
    \end{overpic}
 \vspace{-4mm}
\caption{More shape reconstruction results trained on watertight data. We show four groups of results: the first two rows are reconstructed from 3000 points while the last two rows are generated given 300 points.}
\label{fig:supp_point_complete_car_disn}
\end{figure*}

\begin{figure*}[!htb]
    \centering
    \begin{overpic}[width=\linewidth]{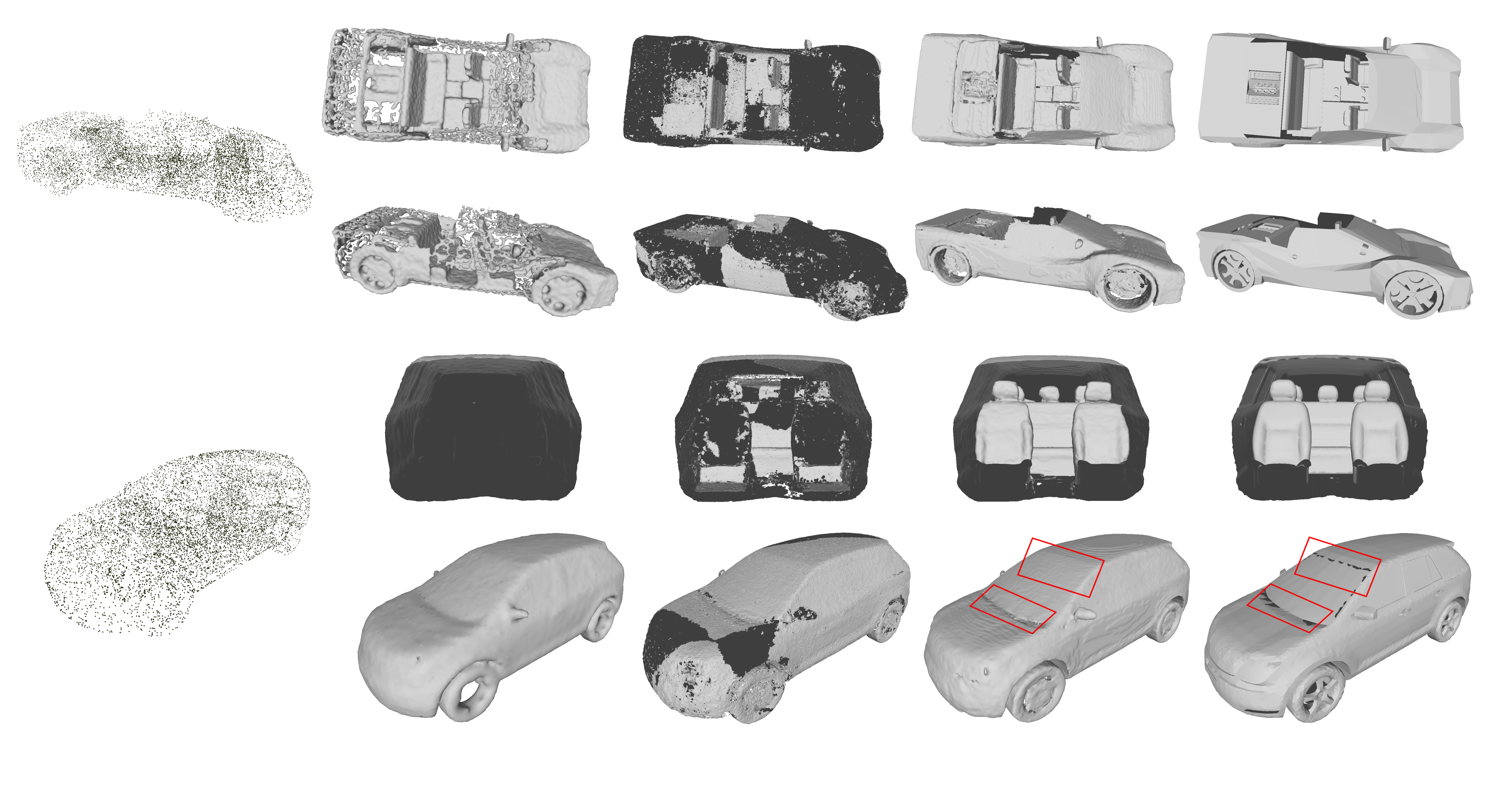}
    \vspace{8mm}
    \put(35, 7) {\small Input}
    \put(145, 7) {\small SAL~\cite{Atzmon_2020_CVPR}}
    \put(240, 7){\small NDF~\cite{chibane2020ndf}}
    \put(340, 7) {\small Ours}
    \put(440, 7) {\small GT}
    \end{overpic}
 \vspace{-4mm}
\caption{More shape reconstruction results trained on unprocessed, raw data. For each group of results, we show the input (10K points) on the left and two rows of corresponding results on the right. 
For the second group of result, we show the inner structure of reconstruction on top of an external view. The highlighted regions within the red rectangles show that our method can generat reconstruction results with consistent normals even when the ground-truth data contain flipped triangles.}
\label{fig:supp_point_complete_car_raw}
\end{figure*}

\begin{figure*}[!htb]
    \centering
    \begin{overpic}[width=\linewidth]{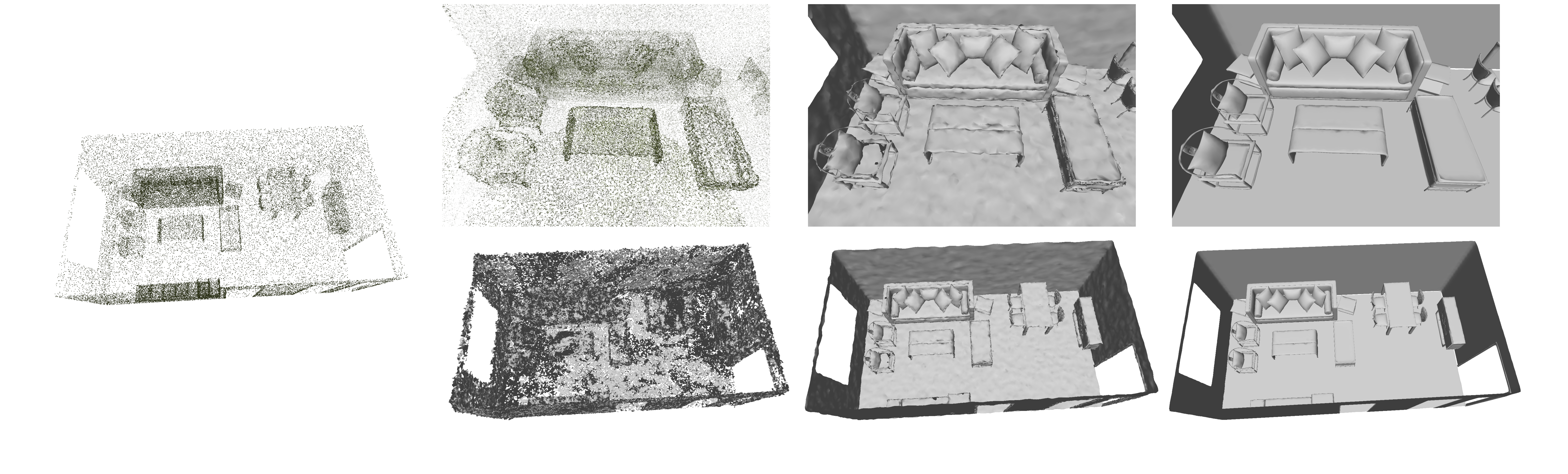}
    \vspace{8mm}
    \put(35, 7) {\small Input}
    \put(180, 7){\small NDF~\cite{chibane2020ndf}}
    \put(300, 7) {\small Ours}
    \put(420, 7) {\small GT}
    \end{overpic}
 \vspace{-4mm}
\caption{Scene reconstruction results from sparse point cloud. For each method, we show both the closeups (first row) and the global view (second row). NDF results contain 3 million points. Note that since we are not able to generate plausible meshing results for NDF even after experimenting with various BPA parameters, we show the output raw point cloud in the closeup of NDF. The other results are displayed in mesh form.}
\label{fig:supp_point_complete_scene}
\end{figure*}

\begin{figure*}[!htb]
    \centering
    \begin{overpic}[width=\linewidth]{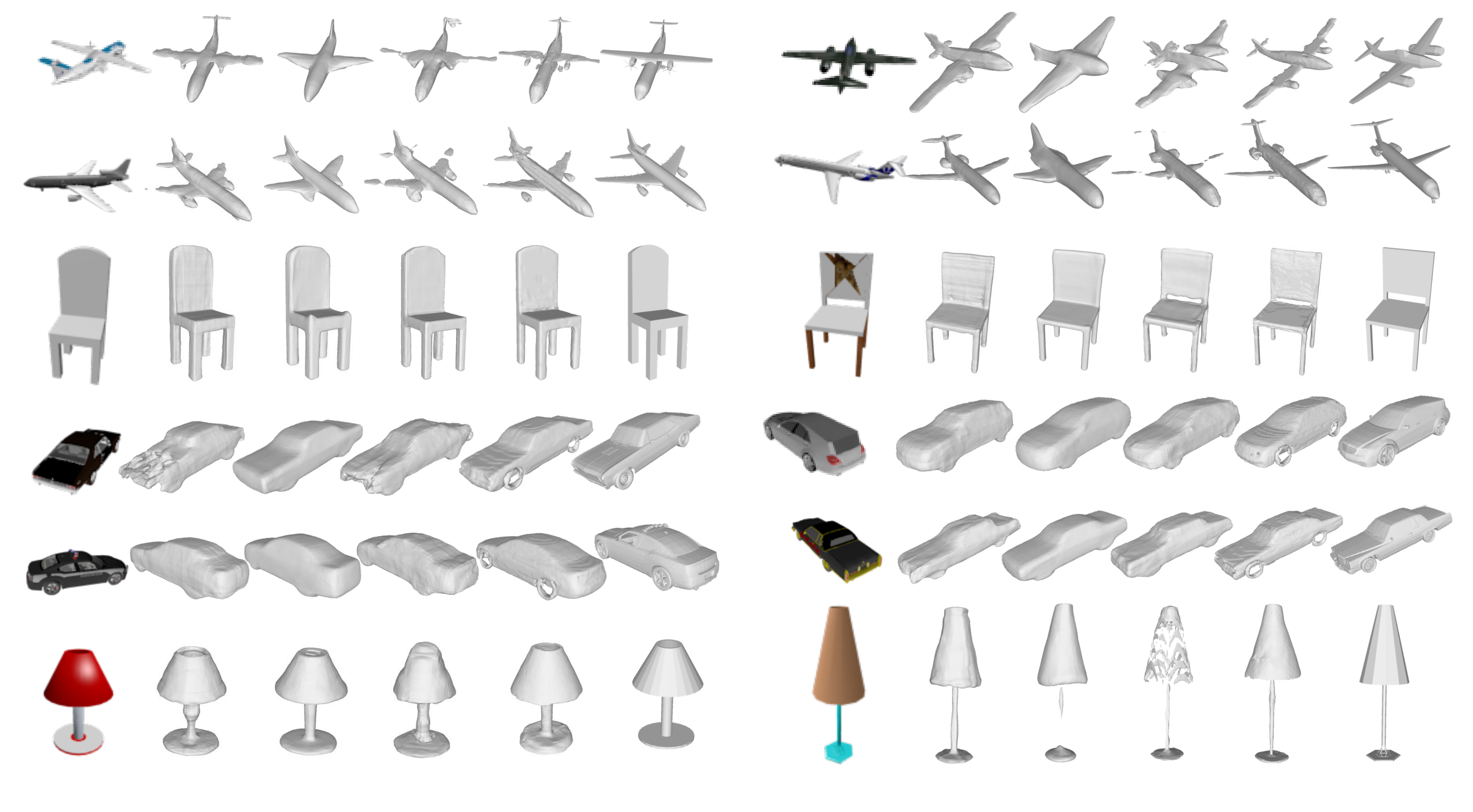}
    \put(15, 7) {\small Input}
    \put(55, 7) {\small IMNet}
    \put(95, 7){\small OccNet}
    \put(135, 7){\small DISN}

    \put(175, 7) {\small Ours}
    \put(220, 7) {\small GT}
    
    \put(275, 7) {\small Input}
    \put(310, 7) {\small IMNet}
    \put(345, 7){\small OccNet}
    \put(385, 7){\small DISN}
    \put(420, 7) {\small Ours}
    \put(460, 7) {\small GT}
    
    \end{overpic}
    
\vspace{-4mm}
\caption{More qualitative comparison results with SOTA single-view reconstruction methods based on implicit functions.}
\label{fig:svr_shapenet_supp}
\vspace{-4mm}
\end{figure*}

\paragraph{Reconstruction of 3D scenes from sparse point cloud.}
In Figure~\ref{fig:supp_point_complete_scene}, we show more visual comparisons of scene reconstruction results. 
The input point cloud (for both main paper and supplementary material) contains 50K points.
Note that we are not able to generate plausible meshing result for NDF even after experimenting with various parameters of BPA algorithm.
Hence we show the raw output point cloud of NDF in the closeup figure.

\paragraph{Single-view reconstruction.} We include more qualitative comparison results on the test set of ShapeNet in Figure~\ref{fig:svr_shapenet_supp}.

    }
    {}

\end{document}